\documentclass[journal,twoside,web]{ieeecolor}
\usepackage{generic}
\usepackage{cite}
\usepackage{amsmath,amssymb,amsfonts}
\usepackage{graphicx}
\usepackage{textcomp}
\usepackage{latexsym}
\usepackage{float}
\usepackage[ruled,vlined]{algorithm2e}
\usepackage{algorithmic}
\usepackage{threeparttable}
\usepackage{multicol}
\usepackage{multirow}
\usepackage{booktabs} 
\usepackage{makecell} 
\usepackage{bm}
\usepackage{bbm}
\usepackage{bbding}
\usepackage{url}
\usepackage{hyperref}
\usepackage{siunitx}

\def\BibTeX{{\rm B\kern-.05em{\sc i\kern-.025em b}\kern-.08em
    T\kern-.1667em\lower.7ex\hbox{E}\kern-.125emX}}
\begin{document}
\title{From Few to More: Scribble-based Medical Image Segmentation via Masked Context Modeling and Continuous Pseudo Labels}

\author{
Zhisong Wang, Yiwen Ye, Ziyang Chen, Minglei Shu, Yanning Zhang, ~\IEEEmembership{Fellow,~IEEE}, and Yong Xia, \IEEEmembership{Member, IEEE}
\thanks{
This work was supported in part by the National Natural Science Foundation of China under Grant 92470101 and Grant 62171377, in part by the "Pioneer" and "Leading Goose" R\&D Program of Zhejiang, China, under Grant 2025C01201(SD2), and in part by the Innovation Foundation for Doctor Dissertation of Northwestern Polytechnical University under Grant CX2024016. (Z. Wang and Y. Ye contributed equally to this work.) (Corresponding author: Y. Xia.)}
\thanks{
Z. Wang, Y. Ye, Z. Chen, and Y. Zhang are with the National Engineering Laboratory for Integrated Aero-Space-Ground-Ocean Big Data Application Technology, School of Computer Science and Engineering, Northwestern Polytechnical University, Xi’an 710072, China (e-mail: {zswang, ywye, zychen}@mail.nwpu.edu.cn, ynzhang@nwpu.edu.cn).}
\thanks{
M. Shu is with the Shandong Artificial Intelligence Institute, Qilu University of Technology (Shandong Academy of Sciences), Jinan 370100, China (e-mail: shuml@sdas.org).}
\thanks{Y. Xia is with the National Engineering Laboratory for Integrated Aero-Space-Ground-Ocean Big Data Application Technology, School of Computer Science and Engineering, Northwestern Polytechnical University, Xi’an 710072, China, with the Ningbo Institute of Northwestern Polytechnical University, Ningbo 315048, China, and also with Research \& Development Institute of Northwestern Polytechnical University in Shenzhen, Shenzhen 518057, China (e-mail: yxia@nwpu.edu.cn).}
}
\maketitle

\begin{abstract}
Scribble-based weakly supervised segmentation methods have shown promising results in medical image segmentation, significantly reducing annotation costs. However, existing approaches often rely on auxiliary tasks to enforce semantic consistency and use hard pseudo labels for supervision, overlooking the unique challenges faced by models trained with sparse annotations. These models must predict pixel-wise segmentation maps from limited data, making it crucial to handle varying levels of annotation richness effectively. In this paper, we propose MaCo, a weakly supervised model designed for medical image segmentation, based on the principle of ``from few to more." MaCo leverages Masked Context Modeling (MCM) and Continuous Pseudo Labels (CPL). MCM employs an attention-based masking strategy to perturb the input image, ensuring that the model’s predictions align with those of the original image. CPL converts scribble annotations into continuous pixel-wise labels by applying an exponential decay function to distance maps, producing confidence maps that represent the likelihood of each pixel belonging to a specific category, rather than relying on hard pseudo labels. We evaluate MaCo on three public datasets, comparing it with other weakly supervised methods. Our results show that MaCo outperforms competing methods across all datasets, establishing a new record in weakly supervised medical image segmentation.

\end{abstract}


\section{Introduction}
\label{sec:introduction}

\IEEEPARstart{M}{edical} image segmentation is critical for modern clinical workflows, supporting preoperative planning, treatment design, and prognostic assessment \cite{azad2024medical}. Although recent advances in convolutional neural networks (CNNs) have enabled accurate and automated segmentation \cite{panayides2020ai, liu2021review}, these fully supervised models depend on large datasets with high-quality, pixel-wise annotations. Building such datasets is a time-consuming and labor-intensive task.

Weakly supervised semantic segmentation (WSSS) has emerged to mitigate this challenge. By leveraging sparse annotations like image-level labels, key points, bounding boxes, and scribbles, WSSS significantly reduces the annotation burden. However, while efficient to acquire, annotations such as image-level labels, key points, and bounding boxes provide limited spatial detail, often leading to suboptimal segmentation accuracy for objects with complex boundaries. In contrast, scribble-based annotations (a few user-drawn lines within target regions) offer a superior balance between cost and information \cite{jin2023label}, providing class-specific guidance more intuitively than bounding boxes \cite{tajbakhsh2020embracing}.

Scribble-based WSSS is dominated by two strategies: consistency learning and label extension. Consistency learning methods \cite{zhou2003learning, luo2021semi} enforce semantic consistency between predictions for original and augmented images.
For instance, ShapePU \cite{zhang2022shapepu} and CycleMix \cite{zhang2022cyclemix} use image masking and mix-augmentation, respectively, to align predictions. Although these methods enhance semantic understanding, they primarily enforce invariance to input perturbations rather than compelling the model to infer detailed structures from sparse cues.
Label extension methods \cite{lee2020scribble2label, zhou2023weakly, luo2022scribble, li2023scribblevc, li2024scribformer}, conversely, propagate scribbles to unlabeled regions to generate hard pseudo labels for denser supervision. This approach, however, is prone to generating inaccurate labels at object boundaries, which propagates errors and degrades model performance.

To address these limitations, we propose \textbf{Ma}sked context modeling and \textbf{Co}ntinuous pseudo labels (MaCo), a novel weakly supervised model operating on the principle of ``\textit{from few to more}." MaCo integrates two synergistic components: Masked Context Modeling (MCM) and Continuous Pseudo Labels (CPL). 
The MCM employs a scribble-guided attention masking strategy, where patches containing scribble pixels are assigned higher masking probabilities while preserving a fixed overall masking ratio. This masking is applied to the original image to generate a masked version.
Both the original and masked images are input into the model, which is trained to predict intact segmentation maps. 
By heavily masking regions surrounding the target object and thus diminishing direct visual cues, the MCM module compels the model to infer semantic structure from sparse signals and surrounding context rather than relying on explicit evidence.
In contrast to methods that focus on semantic consistency, MCM enhances the model's ability to leverage surrounding contextual information, which is essential for WSSS.
Furthermore, CPL generates continuous pseudo labels based on the intuition that pixels closer to scribbles are more likely to belong to the same category. 
Specifically, it computes distance maps by measuring the Euclidean distance from each unlabeled pixel to the nearest scribble pixel, and applies an exponential decay function to transform these distance maps into category-specific confidence maps. These soft labels convey richer semantic information and reflect varying degrees of label confidence, improving supervision fidelity over traditional hard-label approaches.

We evaluated MaCo on three public datasets. Our model not only outperforms existing WSSS methods but also achieves accuracy competitive with fully supervised techniques. Moreover, MaCo demonstrates remarkable robustness to variations in annotation sparsity, maintaining high performance even when trained with fewer samples or sparser scribble pixels. 
Our contributions are three-fold:

\begin{itemize}
 \item We provide a comprehensive analysis of the optimization goals for scribble-based weakly supervised models, culminating in the principle of ``\textit{from few to more}." Based on this insight, we propose the MaCo model for weakly supervised medical image segmentation.
 \item We introduce masked context modeling and continuous pseudo labels to improve the model's ability to predict semantic information from context and enrich supervision signals, respectively.
 \item We demonstrate that MaCo outperforms state-of-the-art scribble-based WSSS methods across three public datasets, maintaining high performance even with reduced training samples or scribble pixels.
\end{itemize}

\section{Related Work}

\begin{figure*}[ht]
    \centering
    \includegraphics[width=0.9\linewidth]{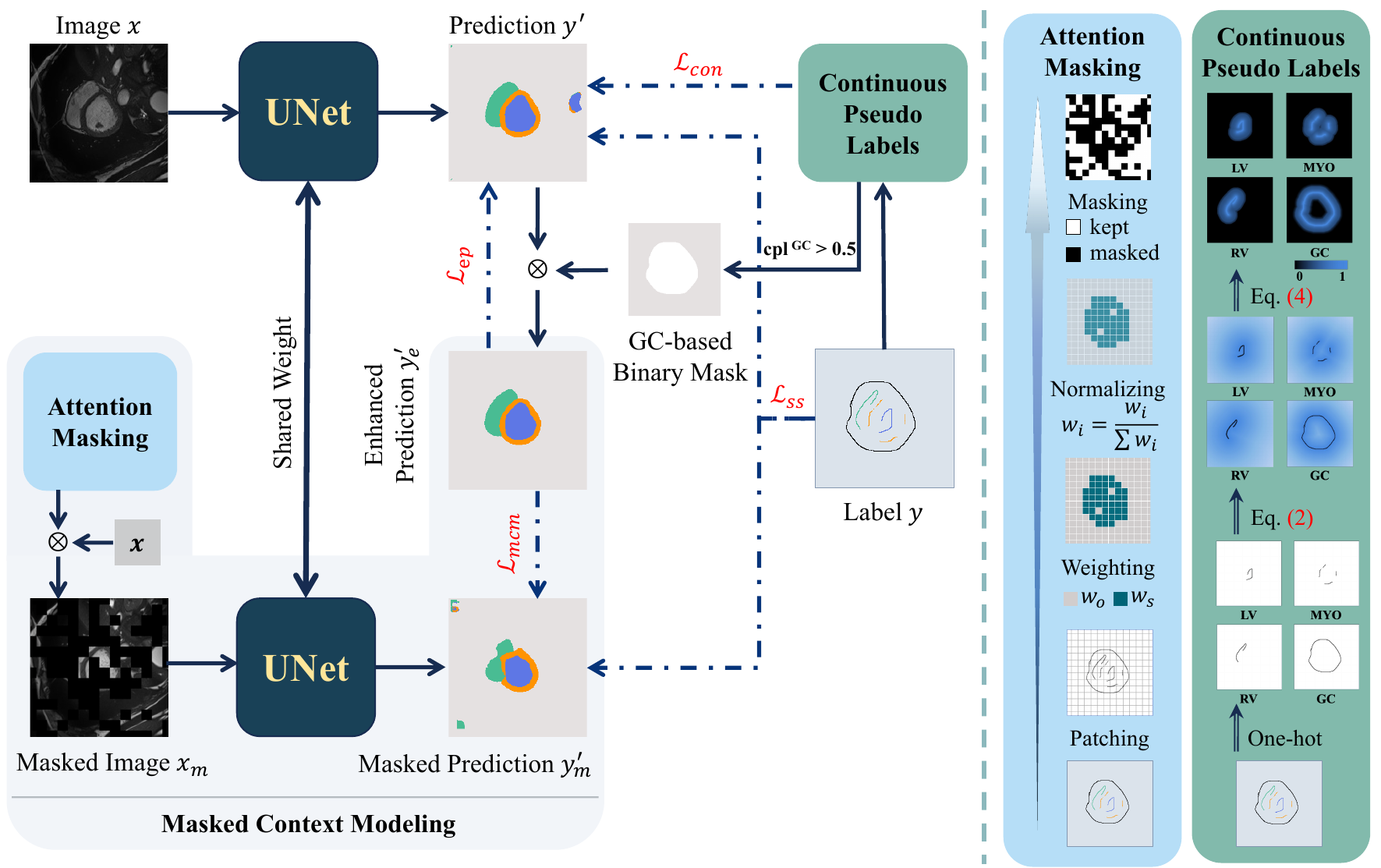}
    \caption{Overview of our MaCo model. This model utilizes MCM and CPL. MCM ensures the model produces consistent predictions for both the original image and its masked version. CPL generates continuous pseudo labels that provide rich supervisory signals while minimizing the risk of introducing misleading information.}
    \label{fig:overview}
\end{figure*}

\subsection{Medical Image WSSS}
The high cost of generating dense annotations has driven significant research into WSSS for medical imaging. WSSS methods aim to train robust segmentation models using various forms of sparse supervision \cite{shen2023survey}.

Common types of weak labels include image-level, point-level, and bounding-box annotations. 
\textit{Image-level supervision} provides a single class label for an entire image \cite{han2022multi, tang2024hunting, zhao2024sfc}. Methods using these labels typically generate Class Activation Maps (CAMs) from an auxiliary classification task to create coarse initial pseudo labels, which are then refined for segmentation training.
\textit{Point-level annotations} mark specific locations, such as extreme points of an object \cite{roth2021going, zhong2023simple} or interior points of a target region\cite{en2022annotation}. Roth \textit{et al.} \cite{roth2021going} utilized a random walk algorithm based on extreme points to produce pseudo labels. The PSCV method \cite{en2022annotation} employs interior points along with a contrastive variance strategy to enhance model performance by comparing variance distributions across different classes.
\textit{Bounding-box annotations} enclose the target object in a rectangle, requiring the model to distinguish foreground from background within the box using techniques like similarity analysis \cite{wei2023weakpolyp} and multi-instance learning \cite{wang2021bounding}.
While these annotation types reduce labeling costs, they provide limited information regarding precise object boundaries. Consequently, models trained on them often struggle to achieve the segmentation accuracy required for complex medical cases.

\textit{Scribble-based supervision} has emerged as a compelling alternative, offering a superior trade-off between annotation efficiency and information richness. Scribbles provide explicit, albeit sparse, pixel-level guidance for each class. Current scribble-based methods can be broadly categorized into two main strategies: consistency learning and label extension.
Consistency learning methods \cite{zhang2022shapepu, liu2022weakly, zhang2022cyclemix} train models to produce consistent predictions for an original image and its randomly augmented version, encouraging the model to learn intrinsic image features and improve robustness. For instance, ShapePU \cite{zhang2022shapepu} applies cutout operations and forces the model to predict masked regions as background, while CycleMix \cite{zhang2022cyclemix} employs a mix-based augmentation strategy and cycle-consistency loss to enforce semantic constraints.
Label extension methods \cite{lee2020scribble2label, luo2022scribble, zhou2023weakly, li2023scribblevc, li2024scribformer} aims to enrich the sparse supervision signal by generating pseudo labels for unlabeled regions. For instance, S2L \cite{lee2020scribble2label} extends labels based on prediction confidence from a single network branch, SC-Net \cite{zhou2023weakly} propagates scribble annotations into unlabeled regions using superpixel-guided scribble-walking, DMPLS \cite{luo2022scribble} employs dual CNN-based decoders to dynamically combine predictions and generate pseudo labels, and ScribFormer \cite{li2024scribformer} integrates a CNN-Transformer architecture to generate higher-quality pseudo labels.
Beyond these two main categories, nnUNet$_{pL}$ \cite{gotkowski2024embarrassingly} employs partial cross-entropy loss and partial Dice loss to train models with limited annotations.

Despite these advances, existing methods do not fully address the core challenge of WSSS: learning to make dense, accurate predictions from sparse initial signals. To address this gap, our work is guided by the principle of ``\textit{from few to more}," where the model must learn to infer complete semantic information from limited contextual cues. We achieve this with MaCo, which combines masked context modeling and a more robust form of pseudo-labeling to enhance contextual inference and provide reliable, enriched supervision.

\subsection{Masked Image Modeling}

Masked Image Modeling (MIM), initially introduced by Masked Autoencoders (MAE) \cite{he2022masked}, serves as a pretext task for self-supervised learning. In the standard MIM pretext task, a significant portion of an image's patches are randomly masked, and an encoder-decoder model is trained to reconstruct the missing patches from the visible ones. This approach has proven highly effective in transfer learning \cite{cai2022uni4eye, wang2023swinmm, ye2024continual}. Many subsequent studies have applied MIM or its underlying principles to various domains, including image inpainting \cite{wang2023imagen}, image generation \cite{chang2022maskgit, gao2023masked}, face recognition \cite{yuan2022msml}, multi-view learning \cite{ji2023multi}, and action representation learning \cite{abdelfattah2024maskclr}. For instance, Imagen Editor \cite{wang2023imagen} uses an object detector-based masking policy to improve alignment between generated images and corresponding text. MaskGIT \cite{chang2022maskgit} employs cosine scheduling as a masking strategy to enhance image generation quality during the decoding phase. MDT \cite{gao2023masked} integrates masking within the diffusion process to obscure potential image representations during training.

We posit that the fundamental capability required for MIM, \textit{i.e.}, inferring missing content from a limited context, is directly analogous to the central challenge of scribble-based WSSS, where a complete segmentation mask must be inferred from a few labeled pixels. Building on this insight, we developed MCM to explicitly integrate this predictive capability into our model.
However, unlike conventional MIM methods that focus on reconstructing raw pixel values \cite{he2022masked, wang2023swinmm}, our MCM module is designed to predict the semantic category of masked regions. This aligns with our ``\textit{from few to more}” principle.

\section{Methodology}

\subsection{Overview}
Our MaCo is a WSSS model comprising two key components: MCM and CPL.
MaCo employs a modified U-Net \cite{baumgartner2018exploration} as its backbone architecture. During training, both the original image and a masked version, generated via an attention masking operation, are fed into the U-Net, resulting in two respective predictions. 
The MCM loss encourages the prediction from the masked image to closely align with the enhanced prediction of the original image, which has been enhanced using global category scribbles. This compels the model to leverage contextual information, thereby improving its inference capabilities.
Meanwhile, CPL enhances the scribble annotations by applying an exponential decay function to distance maps, thereby generating category-specific continuous pseudo labels and improving the reliability of the annotations.
The overall pipeline of MaCo was illustrated in Fig. \ref{fig:overview}. We now delve into the details of each component.

\subsection{Masked Context Modeling}
Unlike existing consistency learning methods that rely on auxiliary tasks to enhance models' awareness of semantic consistency, MaCo directly encourages the model to leverage contextual information for improved inference. To achieve this, MaCo introduces MCM as an auxiliary task, which consists of an attention masking operation and an MCM loss function.

Given an input image $x$, MCM generates a masked version $x_m$ via a scribble-guided attention masking strategy. The scribble annotation is duplicated and divided into $N$ non-overlapping patches $\{P^s_i\}^N_{i=1}$. Each patch $P^s_i$ is assigned a weight $w_i$ based on whether it contains scribble pixels:
\begin{equation}
w_{i} = 
\begin{cases} 
w_s, &  \text{if } P^s_i \text{ contains scribble pixels} \\
w_o, & otherwise
\end{cases} 
\label{eq: ws}
\end{equation}
with $w_s > w_o$. The weights $\{w_i\}$ are normalized to form a discrete probability distribution $\{w^\prime_i\}$, where $w^\prime_i = \frac{w_i}{\sum_{j=1}^{N} w_j}$. A weighted random sampling procedure is then used to select a fixed proportion $\varphi$ of patches to be masked (set to 0), while the remaining patches are retained (set to 1), resulting in a binary attention mask $A \in \{0,1\}^{H \times W}$.

The masked image is computed as $x_m =x\odot A$, where $\odot$ denotes element-wise multiplication. Both the original and masked images, $x$ and $x_m$, are passed through a shared U-Net backbone to generate predictions $y^{\prime}$ and $y_{m}^{\prime}$, respectively. The full masking procedure is outlined in Algorithm \ref{algorithm_mcm}.

Moreover, the scribble annotations include both foreground scribbles and global category (GC) scribbles. The GC scribbles define the extent of the target region, ensuring all targets are covered, and providing an initial estimate of the foreground area. This information is used to refine the predictions, particularly in the early stages of training. A binary mask based on the GC scribbles is generated using the GC continuous pseudo labels (introduced in Subsection \ref{CPL}). Pixels in the GC continuous pseudo labels with values less than 0.5 are set to 0, while the remaining pixels are set to 1. This GC-based binary mask is then used to produce an enhanced prediction $y_{e}^{\prime}$ that provides more accurate supervision signals for $y_{m}^{\prime}$.
To ensure consistency between the predictions, we introduce a masked context modeling loss $\mathcal{L}_{\text{mcm}}$, defined as:
\begin{equation}
\mathcal{L}_{\text{mcm}} = 1 - \frac{ y_{m}^{\prime} \cdot y_{e}^{\prime} }{\| y_{m}^{\prime} \| \| y_{e}^{\prime} \|},
\end{equation}
where $\| \cdot \|$ denotes the L2-norm.
This loss measures the cosine similarity between the masked prediction $y_{m}^{\prime}$ and the enhanced prediction $y_{e}^{\prime}$, encouraging semantic alignment between them.

By optimizing $\mathcal{L}_{\text{mcm}}$, the model is able to leverage contextual information to infer the semantic content of masked regions, facilitating learning from sparse annotations.

\begin{algorithm}
\caption{Attention Masking Algorithm}
\label{algorithm_mcm}
\LinesNumbered
\SetKwInOut{Input}{Input}
\SetKwInOut{Output}{Output}
\SetKwInOut{Parameter}{Parameter}
\Input{Input image $x$, scribble annotation $S$}
\Parameter{Masking ratio $\varphi$, high masking weight $w_\text{s}$, lower masking weight $w_\text{o}$,}
\Output{Masked image $x_m$}
Partition scribble annotation $S$ into $N$ patches $\{P^s_i\}_{i=1}^N$ \\
Initialize sampling weights $\{w_i\}_{i=1}^N$ for all patches \\
\For{$i = 1$ \KwTo $N$}{
    \eIf{$P^s_i$ contains scribble pixels}{
        $w_i = w_\text{s}$ 
    }{
        $w_i = w_\text{o}$ 
    }
}

Normalize weights: $w^\prime_i = \frac{w_i}{\sum_{j=1}^{N} w_j}$ \\
Compute masked patches count: $K = \lfloor \varphi \times N \rfloor$ \\
Sample $U \subseteq \{1,\dots,N\}$ via weights $\{w^\prime_i\}$, $|U| = K$  \\
\For{$i = 1$ \KwTo $N$}{
    \eIf{$i \in U$}{
        Set $P^s_i = 0$ 
    }{
        Set $P^s_i = 1$ 
    }
}
Reconstruct attention mask $A$ by assembling $\{P^s_i\}_{i=1}^N$ \\
Compute masked image: $x_m = x \odot A$ \\
\Return $x_m$
\end{algorithm}

\subsection{Continuous Pseudo Labels}
\label{CPL}
To address the limitations of hard pseudo labels, which can lead to inaccurate predictions and misguide the model, we propose the CPL component based on distance priors. CPL aims to balance rich supervision signals with potentially misleading information by generating continuous pseudo labels.

\begin{figure}[ht]
    \centering
    \includegraphics[width=0.85\linewidth]{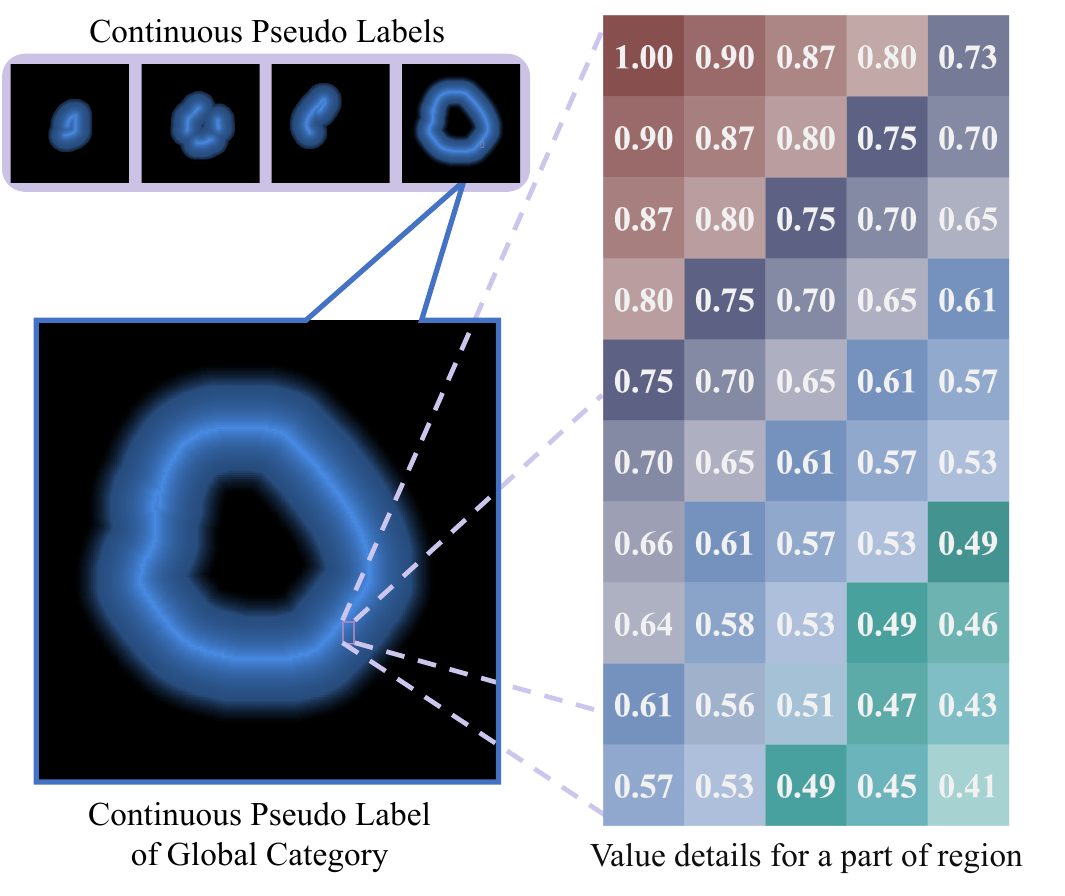}
    \caption{Visualization of the global category continuous pseudo label. The purple boxes indicate the value details within a selected region, showing that the value decreases as the distance from the scribble increases.}
    \label{fig:continue}
\end{figure}

CPL starts by splitting the scribble annotation into multiple binary annotations, each representing a specific category. 
For each category $c$, let $S^c$ denote the set of annotated pixels. The Euclidean distance from every image pixel $p = (i, j)$ to the nearest scribble pixel $(i_s, j_s)$ is computed to obtain the distance map $D^c(p)$ :
\begin{equation}
D^c(p) = \min_{(i_s, j_s) \in S^c} \sqrt{(i - i_s)^2 + (j - j_s)^2}.
\end{equation}
This distance $D^c(p)$ is then converted into a continuous value using an exponential decay function, yielding the continuous pseudo label $cpl^c$, defined as:
\begin{equation}
cpl^c = \{cpl^c_{p}\},
\end{equation}
\begin{equation}
cpl^c_{p} = 
\begin{cases} 
e^{-0.1 \cdot D^c(p)} \cdot \mathbb{I}(e^{-0.1 \cdot D^c(p)} > 0.05), &  \text{if } c \neq \text{GC} \\
\max(e^{-0.1 \cdot D^c(p)},  0.05), & \text{if } c = \text{GC}
\end{cases}
\label{eq: cpl}
\end{equation}
where $\mathbb{I}(\cdot)$ denotes the indicator function. 
The full procedure is detailed in Algorithm~\ref{algorithm_cpl}. A visual example of the resulting $cpl^c$ is depicted in Fig.~\ref{fig:continue} to illustrate its spatial structure.

In our CPL formulation, pixels closer to a scribble are assigned higher confidence scores, ranging from 0 to 1. We set the lower confidence limit to 0.05, as the prior information from scribble pixels becomes increasingly unreliable at greater distances. 
Specifically, global category scribbles (also known as background scribbles) are manually-drawn annotations that precisely outline background regions in images. Within the MaCo framework, these scribbles serve dual purposes: (1) the CPL module utilizes them through a confidence-weighted mechanism to sharpen boundary perception for target segmentation, and (2) during prediction enhancement, they guide spatial attention to foreground targets while suppressing background regions. Specifically, we employ a threshold of 0.5 to filter out peripheral areas in cpl$^{GC}$, thereby compelling the model to focus more intensively on the target regions.

The resulting continuous pseudo labels are then used as supervision signals for $y^\prime$ through the proposed loss function ${L}_{\text{con}}$, formulated as:
\begin{equation}
\mathcal{L}_{\text{con}} = - \frac{1}{|C||N|}  \sum_{c=1}^{C} cpl^c \cdot {y^\prime}_c \log {y^\prime}_c,
\label{loss_con}
\end{equation}
where $C$ is the number of categories, $N= \| cpl^c \|_0$ is the number of non-zero elements in $cpl^c$, $y^\prime_c$ is the model's predicted probability for category $c$, and $\text{cpl}^c$ is the corresponding continuous pseudo-label. This loss functions as a cross-entropy objective where $cpl^c \cdot {y^\prime}_c$ serves as a confidence-weighted supervision signal, ensuring that the model learns more reliably from pixels closer to the original scribbles.

\begin{algorithm}[t]
\caption{CPL Algorithm}
\label{algorithm_cpl}
\LinesNumbered
\SetKwInOut{Input}{Input}
\SetKwInOut{Output}{Output}
\SetKwInOut{Parameter}{Parameter}

\Input{Scribble annotation $S$ with $C$ categories}
\Parameter{Decay factor $\alpha$, confidence threshold $\tau$}
\Output{Continuous pseudo labels $\{\mathrm{cpl}^c\}_{c=1}^{C}$}

\ForEach{category $c = 1$ \KwTo $C$}{
    Extract class-specific scribble pixels: $S^c $ \\

    \ForEach{every image pixel $p = (i,j)$}{
        \ForEach{scribble point $s = (i_s,j_s) \in S^c$}{
            Compute the minimum Euclidean distance: $D^c(p) = \min_{(i_s, j_s)}\sqrt{(i - i_s)^2 + (j - j_s)^2}$ }
        
        Compute exponential decay: $\delta = e^{-\alpha \cdot D^c(p)}$ \\
        \eIf{$c = \mathrm{GC}$}{
            $\mathrm{cpl}^c(p) = \max(\delta, \tau)$
        }{
            $\mathrm{cpl}^c(p) = \delta \cdot \mathbb{I}(\delta > \tau)$ }}}
\Return $\{\mathrm{cpl}^c\}_{c=1}^C$ 
\end{algorithm}

\begin{table*}[ht]
  \centering
  \small
  \setlength\tabcolsep{6pt}     
  \renewcommand{\arraystretch}{1.2} 
  \caption{Results of MaCo, 11 scribble-based methods, and three fully supervised methods across three datasets. `$-$' means training masks are unavailable. `Avg.' is the average over foreground categories. Best in each column is \textbf{bold}.}

  \resizebox{\textwidth}{!}{  
    \begin{tabular}{l  c  *{4}{c}  *{4}{c}  *{3}{c} }
      \toprule
      \multirow{2}{*}{Method} & \multirow{2}{*}{Supervision}
        & \multicolumn{4}{c}{ACDC} 
        & \multicolumn{4}{c}{MSCMRseg} 
        & \multicolumn{3}{c}{NCI-ISBI} \\
      \cmidrule(lr){3-6}\cmidrule(lr){7-10}\cmidrule(l){11-13}
        &  & LV & MYO & RV & Avg. & LV & MYO & RV & Avg. & PZ & CG & Avg. \\
      \midrule
      UNet \cite{ronneberger2015u}       & masks & 89.2 & 83.0 & 78.9 & 83.7 & 85.0 & 72.1 & 73.8 & 77.0 & 72.5 & 82.2 & 77.4 \\
      CycleMix$_F$ \cite{zhang2022cyclemix} & masks & 91.9 & 85.8 & 88.2 & 88.6 & 86.4 & 78.5 & 78.1 & 81.0 & 73.7 & 84.7 & 79.2 \\
      
      Swin-UNet  \cite{cao2022swin}         & masks & 92.2 & 88.2 & 86.3 & 88.9 & $-$ & $-$ & $-$ & $-$ & 72.3 & 84.6 & 78.5 \\
      Mamba-UNet  \cite{wang2024mamba}      & masks & 92.4 & 88.3 & 87.1 & 89.3 & $-$ & $-$ & $-$ & $-$ & 72.2 & 86.5 & 79.4 \\
      TransUNet  \cite{chen2024transunet}      & masks & 92.5 & 88.5 & 89.0 & 90.0 & $-$ & $-$ & $-$ & $-$ & 74.0 & 85.7 & 79.9 \\
  
      MaCo$_F$                      & masks & 94.2 & 90.9 & 90.4 & 91.8 & $-$ & $-$ & $-$ & $-$ & 77.0 & 87.8 & 82.4 \\
      \midrule
      UNet$^+_{pCE}$          & scribbles & 79.9 & 80.1 & 68.5 & 76.2 & 73.4 & 64.0 & 49.7 & 62.4 & 13.2 & 26.1 & 19.7 \\
      UNet$_{RW}$ \cite{grady2006random}  & scribbles & 85.6 & 71.4 & 80.3 & 79.1 & 81.9 & 69.3 & 76.6 & 75.9 & 66.4 & 74.7 & 70.6 \\
      USTM \cite{liu2022weakly}             & scribbles & 82.4 & 80.0 & 82.2 & 81.5 & 78.9 & 67.6 & 67.6 & 71.4 & 52.2 & 65.4 & 58.8 \\
      DMPLS \cite{luo2022scribble}          & scribbles & 91.7 & 84.6 & 86.9 & 87.7 & 85.9 & 76.9 & 83.7 & 82.2 & 30.3 & 83.4 & 56.9 \\
      SC-Net \cite{zhou2023weakly}          & scribbles & 89.0 & 85.1 & 87.6 & 87.2 & 88.3 & 80.8 & 86.9 & 85.3 & 22.6 & 80.9 & 51.8 \\
      S2L \cite{lee2020scribble2label}      & scribbles & 80.9 & 82.3 & 83.6 & 82.3 & 81.7 & 81.1 & 72.1 & 78.3 & 72.3 & 63.9 & 68.1 \\
      CycleMix$_S$ \cite{zhang2022cyclemix} & scribbles & 88.3 & 79.8 & 86.3 & 84.8 & 87.0 & 73.9 & 79.1 & 80.0 & 66.4 & 75.7 & 71.1 \\
      ScribFormer \cite{li2024scribformer}  & scribbles & 92.2 & 87.1 & 87.1 & 88.8 & 89.6 & 81.3 & 80.7 & 83.9 & 63.5 & 77.4 & 70.5 \\
      nnUNet$_{pL}$ \cite{gotkowski2024embarrassingly} & scribbles & 84.2 & 84.3 & 82.9 & 83.8 & 89.5 & \textbf{85.4} & 87.0 & 87.3 & 65.1 & 83.1 & 74.1 \\
      ShapePU \cite{zhang2022shapepu}       & scribbles & 86.0 & 79.1 & 85.2 & 83.4 & 91.9 & 83.2 & 80.4 & 85.2 & 71.6 & 83.2 & 77.4 \\
      ScribbleVC \cite{li2023scribblevc}    & scribbles & 91.4 & 86.6 & 87.0 & 88.4 & 92.1 & 83.0 & 85.2 & 86.8 & 70.5 & 81.6 & 76.1 \\
      \midrule
      \textbf{MaCo}                           & scribbles & \textbf{93.4} & \textbf{89.2} & \textbf{88.7} & \textbf{90.4} & \textbf{93.1} & 84.7 & \textbf{87.8} & \textbf{88.5} & \textbf{73.8} & \textbf{87.1} & \textbf{80.5} \\
      \bottomrule
    \end{tabular}
  }
  \label{tab:three data}
\end{table*}

\begin{table}[t]
  \centering
  \Large                        
  \setlength\tabcolsep{2pt}      
  \renewcommand{\arraystretch}{1.2}
  \caption{Comparative Performance Analysis of the CPL Module Using Different Decay Functions. Best in each column is \textbf{bold}.}
  \resizebox{\linewidth}{!}{%
    \begin{tabular}{l  cccc  cccc  ccc}
      \toprule
      \multirow{2}{*}{\shortstack{Decay\\Function}}
        & \multicolumn{4}{c}{ACDC} 
        & \multicolumn{4}{c}{MSCMRseg} 
        & \multicolumn{3}{c}{NCI-ISBI} \\
      \cmidrule(lr){2-5}\cmidrule(lr){6-9}\cmidrule(l){10-12}
        & LV & MYO & RV & Avg.
        & LV & MYO & RV & Avg.
        & PZ & CG & Avg. \\
      \midrule
        Linear            
        & 92.1 & 88.4 & 87.4 & 89.3 
        & 92.2 & 83.5 & 87.5 & 87.7
        & 63.3 & 80.2 & 71.8 \\
        Gaussian            
        & 92.3 & 88.8 & 87.7 & 89.6
        & 92.6 & 84.1 & 87.6 & 88.1 
        & 72.4 & 83.1 & 77.8 \\
        \midrule
        \textbf{Ours}            
        & \textbf{93.4} & \textbf{89.2} & \textbf{88.7} & \textbf{90.4} 
        & \textbf{93.1} & \textbf{84.7} & \textbf{87.7} & \textbf{88.5}
        & \textbf{73.8} & \textbf{87.1} & \textbf{80.5} \\
      \bottomrule
    \end{tabular}%
  }
  \label{tab:three_delay}
\end{table}

\begin{table*}[ht]
  \centering
  \scriptsize
  \setlength\tabcolsep{6pt}       
  \renewcommand{\arraystretch}{1.2} 
  \caption{Ablation studies of $\mathcal{L}_{\text{pCE}}$, $\mathcal{L}_{\text{mcm}}$, $\mathcal{L}_{\text{mpCE}}$, $\mathcal{L}_{\text{ep}}$, and $\mathcal{L}_{\text{con}}$ on three datasets. Best in each column is \textbf{bold}.}
  \label{tab:ablation_three}
  \resizebox{\textwidth}{!}{%
    \begin{tabular}{ccccc *{4}{c} *{4}{c} *{3}{c}}
      \toprule
      \multirow{2}{*}{$\mathcal{L}_{\text{pCE}}$} 
        & \multirow{2}{*}{$\mathcal{L}_{\text{mcm}}$} 
        & \multirow{2}{*}{$\mathcal{L}_{\text{mpCE}}$} 
        & \multirow{2}{*}{$\mathcal{L}_{\text{ep}}$} 
        & \multirow{2}{*}{$\mathcal{L}_{\text{con}}$} 
        & \multicolumn{4}{c}{ACDC} 
        & \multicolumn{4}{c}{MSCMRseg} 
        & \multicolumn{3}{c}{NCI-ISBI} \\
      \cmidrule(lr){6-9}\cmidrule(lr){10-13}\cmidrule(l){14-16}
      & & & & 
        & LV & MYO & RV & Avg. 
        & LV & MYO & RV & Avg. 
        & PZ & CG & Avg. \\
      \midrule
      $\checkmark$ &           &           &           &           & 79.9 & 80.1 & 68.5 & 76.2 & 73.4 & 64.0 & 49.7 & 62.4 & 13.2 & 26.1 & 19.7 \\
      $\checkmark$ & $\checkmark$ &           &           &           & 90.4 & 88.0 & 85.6 & 88.0 & 88.6 & 73.7 & 79.3 & 80.5 & 52.0 & 66.3 & 59.2 \\
      $\checkmark$ & $\checkmark$ & $\checkmark$ &           &           & 92.0 & 88.1 & 86.2 & 88.8 & 91.3 & 82.6 & 86.8 & 86.9 & 64.4 & 73.3 & 68.9 \\
      $\checkmark$ & $\checkmark$ & $\checkmark$ & $\checkmark$ &           & 92.3 & 88.6 & 87.8 & 89.6 & 92.2 & 83.9 & 86.7 & 87.6 & 66.2 & 83.1 & 74.7 \\
      $\checkmark$ &           &           &           & $\checkmark$ & 89.2 & 85.0 & 79.4 & 84.5 & 82.8 & 74.4 & 73.6 & 76.9 & 28.2 & 57.7 & 43.0 \\
      $\checkmark$ & $\checkmark$ & $\checkmark$ & $\checkmark$ & $\checkmark$ 
        & \textbf{93.4} & \textbf{89.2} & \textbf{88.7} & \textbf{90.4} 
        & \textbf{93.1} & \textbf{84.7} & \textbf{87.8} & \textbf{88.5} 
        & \textbf{73.8} & \textbf{87.1} & \textbf{80.5} \\
      \bottomrule
    \end{tabular}%
  }
\end{table*}

\subsection{Scribble-based Supervision}
Additionally, we incorporate a partial Cross-Entropy (pCE) function \cite{tang2018normalized}, which applies cross-entropy loss exclusively to the pixels labeled as foreground in the scribble annotations. This is used to compute the scribble supervision loss function $\mathcal{L}_{\text{ss}}$, defined  as:
\begin{equation}
\mathcal{L}_{\text{ss}}(y^{\prime}, y_{m}^{\prime}) = \mathcal{L}_{\text{pCE}}(y^{\prime}) + \lambda_{\text{1}}\mathcal{L}_{\text{pCE}}(y_{m}^{\prime}),
\end{equation}
where $\lambda_{\text{1}}$ is a weighting factor.

Similar to $\mathcal{L}_{\text{mcm}}$, we also employ $\mathcal{L}_{\text{ep}}$ to align the prediction $y^{\prime}$ with the enhanced prediction $y^{\prime}_{e}$, thereby improving the model's ability to learn the background category. The enhanced prediction loss $\mathcal{L}_{\text{ep}}$ is defined as:
\begin{equation}
\mathcal{L}_{\text{ep}} = 1 - \frac{ y^{\prime} \cdot y_{e}^{\prime} }{\| y^{\prime} \| \| y_{e}^{\prime} \|}.
\end{equation}

The total loss function combines all individual losses and is defined as:
\begin{equation}
\mathcal{L}_{\text{total}} = \mathcal{L}_{\text{ss}} + \lambda_{\text{2}} \mathcal{L}_{\text{mcm}} + \lambda_{\text{3}} \mathcal{L}_{\text{ep}} + \lambda_{\text{4}} \mathcal{L}_{\text{con}},
\end{equation}
where $\lambda_{\text{2}}$, $\lambda_{\text{3}}$, and $\lambda_{\text{4}}$ are weighting factors.

\section{Experiments}

\subsection{Datasets}
We evaluated the performance of our MaCo model using three scribble-based datasets. 

\subsubsection{ACDC dataset}
This dataset contains cine-MRI scans from 150 subjects \cite{bernard2018deep}. Scribble annotations, available for scans from 100 subjects \cite{valvano2021learning}, label three categories: left ventricle (LV), right ventricle (RV), myocardium (MYO), and a global category (GC). These scribble annotations were used exclusively during training and were not provided during testing. We followed \cite{li2024scribformer} to partition the dataset into training, validation, and test sets in a 70:15:15 ratio, using samples from half of the subjects for training.

\begin{figure}[ht]
    \centering
    \includegraphics[width=0.95\linewidth]{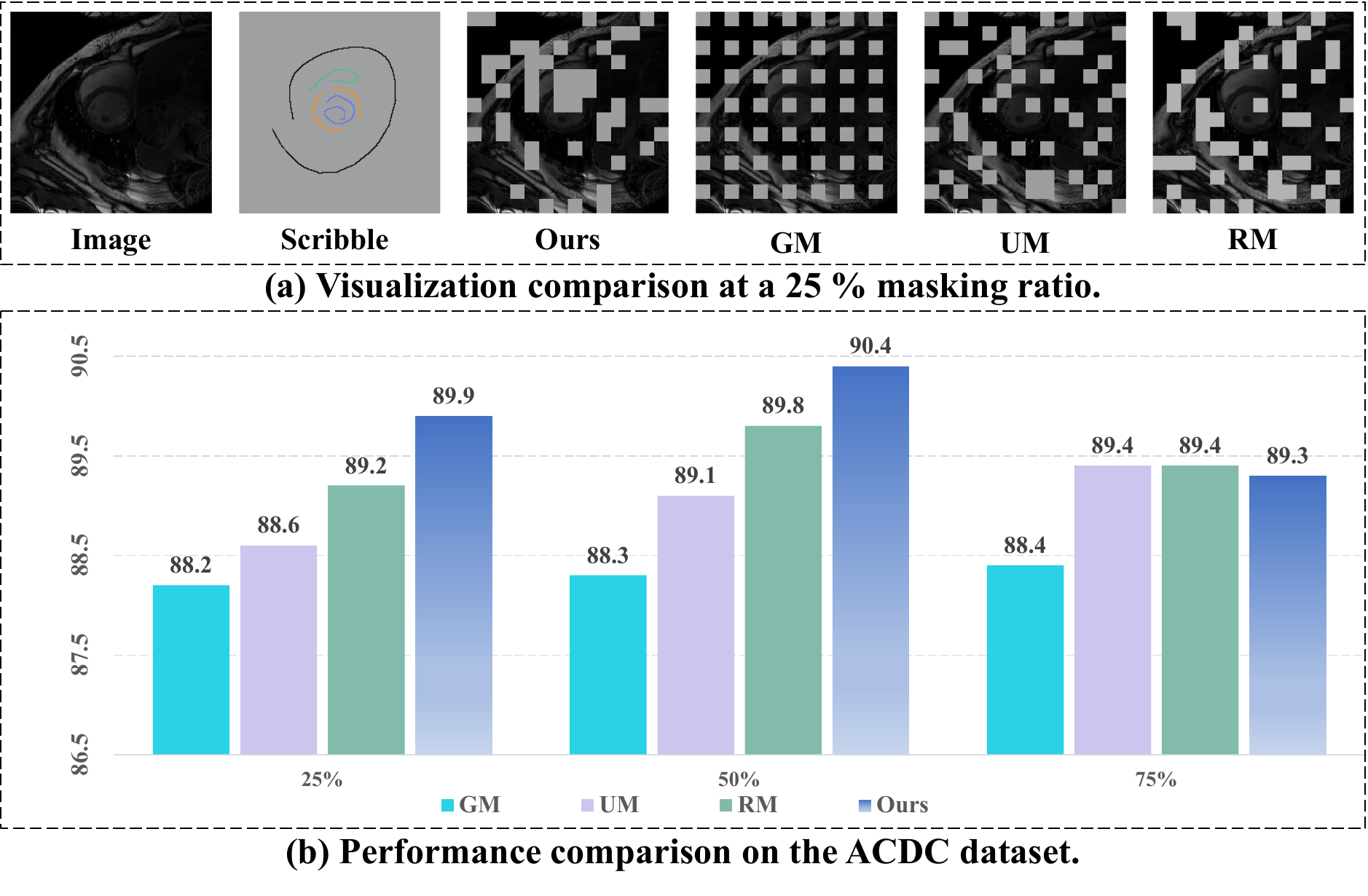}
    \caption{Visualization and performance comparison of four different masking strategies. (a) shows the visualizations of the four masking strategies at a 25\% mask ratio, while (b) presents four strategies segmentation performance on the ACDC dataset. Here, GM denotes Grid Mask, UM denotes Uniform Mask, and RM denotes Random Mask.}
    \label{fig:four_mask}
\end{figure}

\subsubsection{MSCMRseg dataset}
This dataset includes late gadolinium enhancement (LGE) MRI scans from 45 subjects with cardiomyopathy, annotated for LV, RV, MYO, and GC\cite{zhuang2018multivariate}. Scribble annotation data was obtained from \cite{zhang2022shapepu}. Following \cite{zhang2022cyclemix}, we split the dataset into 25 scans for training, 5 for validation, and 15 for testing.

\subsubsection{NCI-ISBI dataset}
This dataset consists of 80 T2-weighted MRI scans from the ISBI 2013 Prostate MRI Challenge \cite{clark2013cancer}. Scribble annotations, provided by \cite{luo2022scribble}, cover central glands (CG), peripheral zones (PZ), and GC. We randomly divided the dataset into 50 scans for training, 15 for validation, and 15 for testing.

\subsection{Implementation Details}

\begin{figure}[ht]
    \centering
    \includegraphics[width=\linewidth]{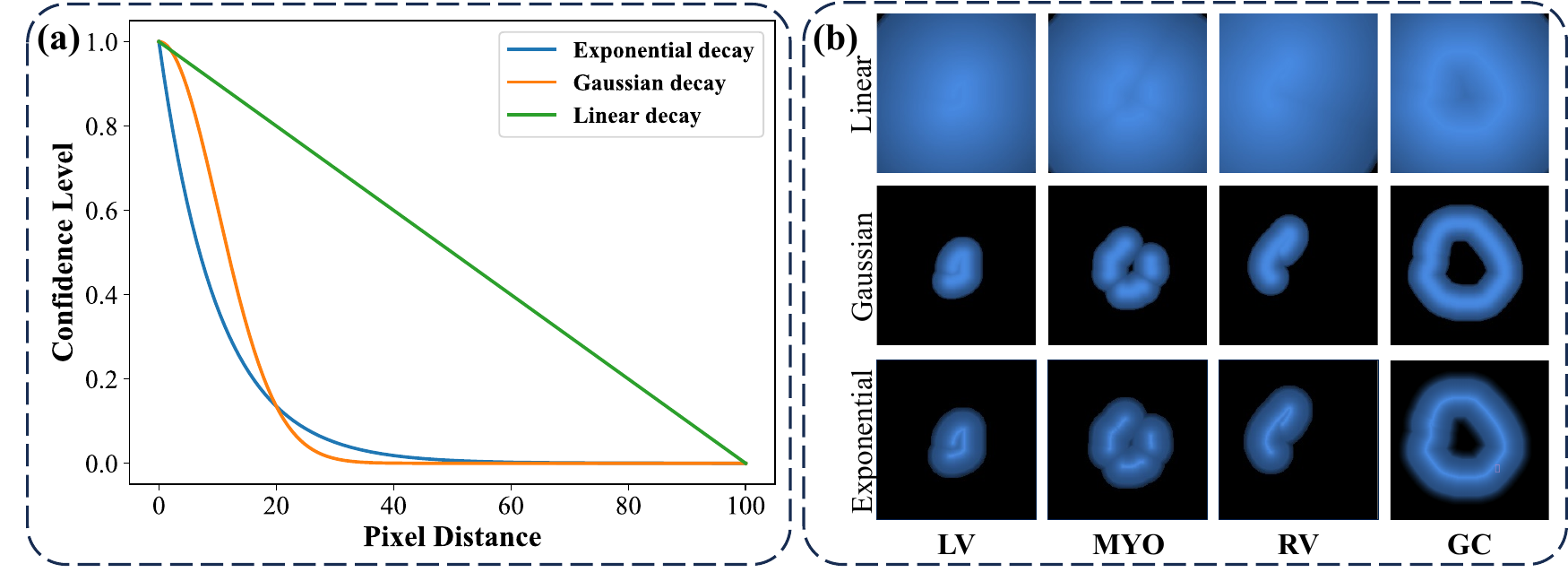}
    \caption{Visualization comparison of three different decay functions in CPL, where (a) shows the comparison of their decay function curves, and (b) is the visualization of three confidence levels of a case on the ACDC dataset.}
    \label{fig:three_delay}
\end{figure}

\begin{figure}[ht]
    \centering
    \includegraphics[width=\linewidth]{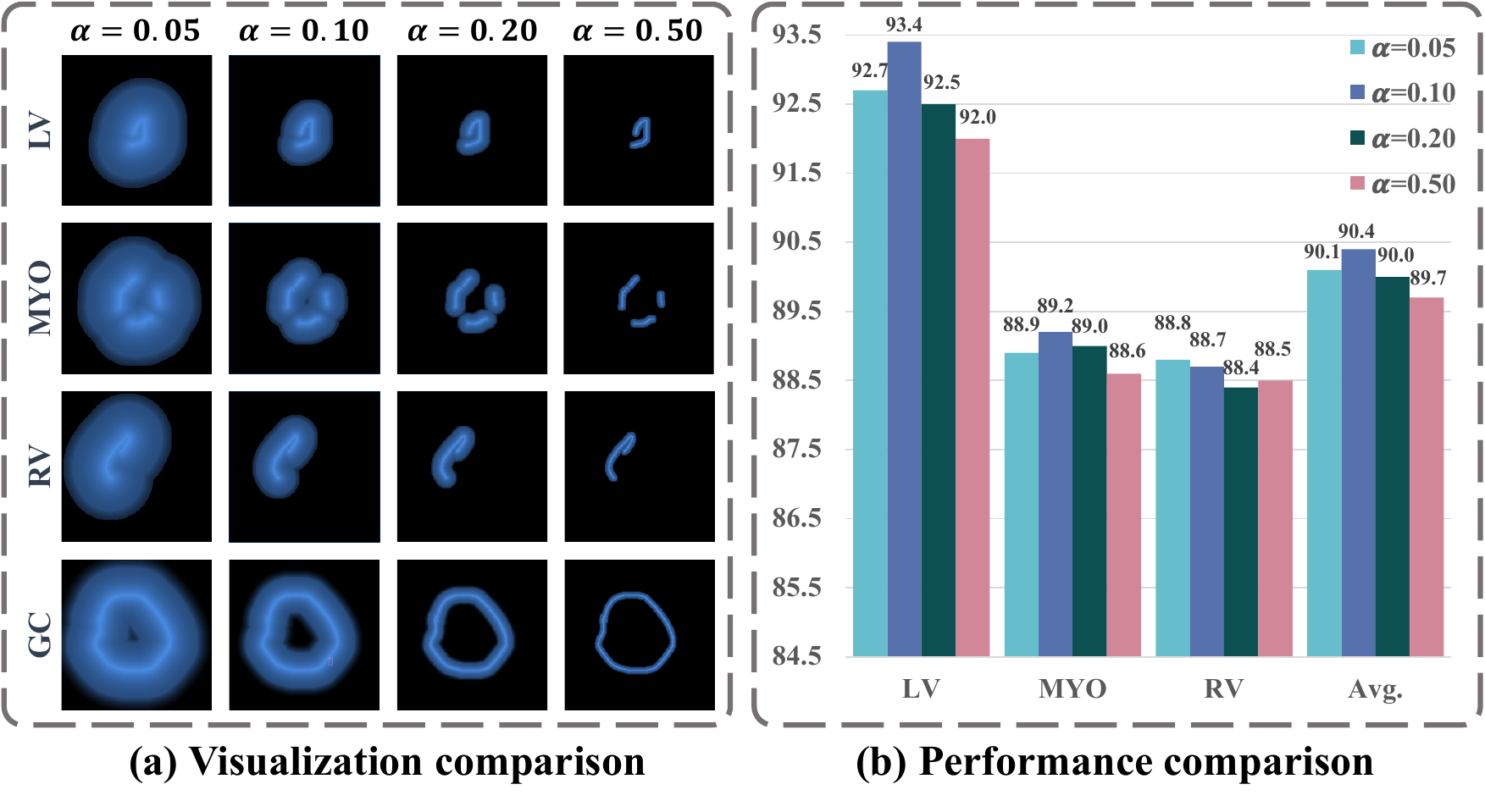}
    \caption{Visualization and performance of CPL under varying decay factors ($\alpha$ = 0.05, 0.1, 0.2, 0.5) on the ACDC dataset. (a) Visual comparison of continuous pseudo labels for LV, MYO, RV, and GC. (b) Quantitative performance (Dice scores) under different decay settings.}
    \label{fig:diff_e}
\end{figure}

\begin{figure}[ht]
    \centering
    \includegraphics[width=\linewidth]{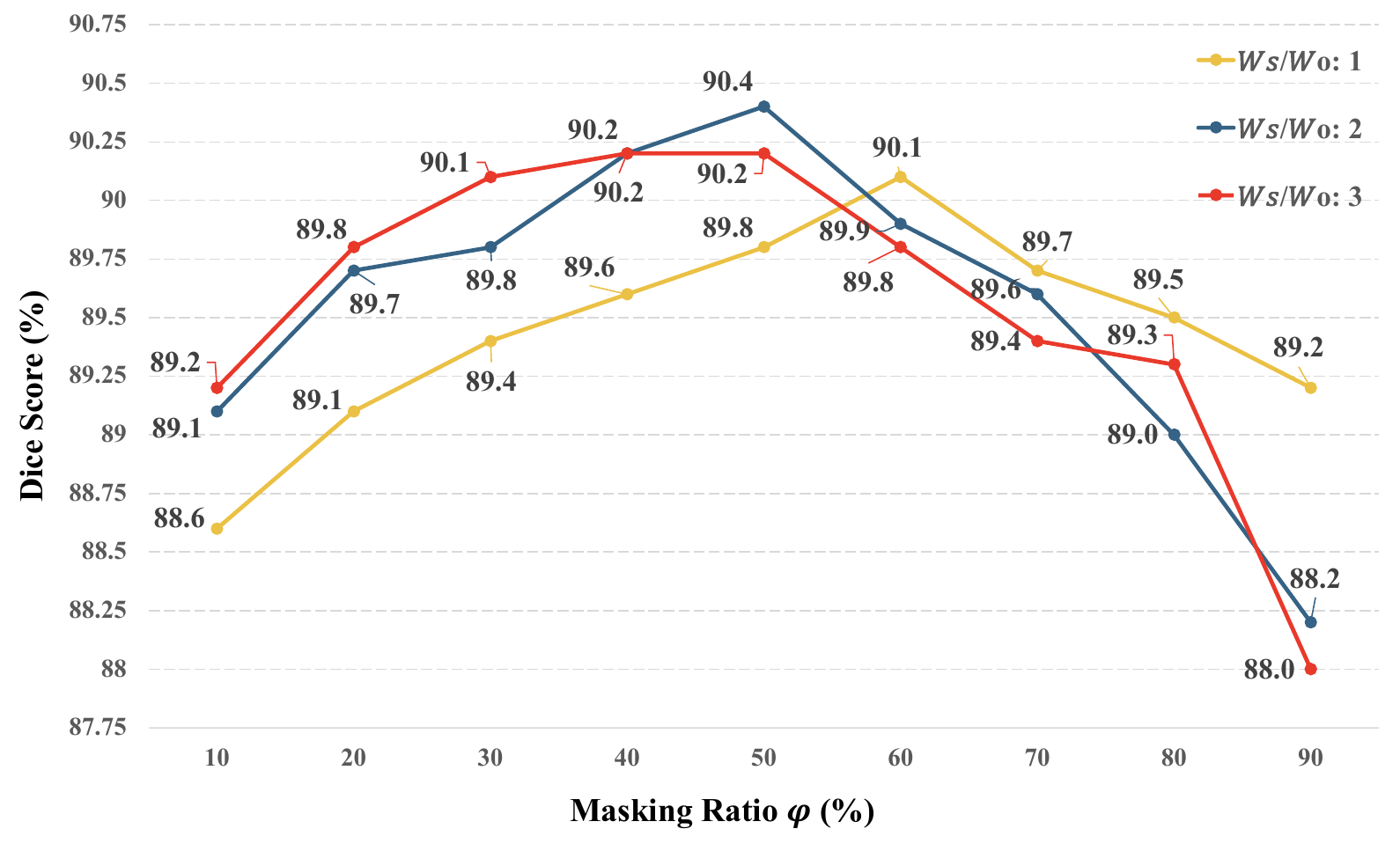}
    \caption{Performance analysis under varying masking weights $w_s$ and masking ratios $\varphi$ on the ACDC dataset. The yellow, blue, and red curves represent $w_s / w_o$ = 1, 2, and 3, respectively.}
    \label{fig:line_chart}
\end{figure}

We utilized an improved 2D UNet$^{+}$ \cite{baumgartner2018exploration} as the backbone for our MaCo model. All images were resampled to an in-plane resolution of $1.37\times1.37 mm^2$ and then cropped to $212\times212$ using a combination of cropping and padding operations. Each resized image was normalized to have zero mean and unit variance. Training was conducted for 300 epochs on the ACDC dataset and 1000 epochs on the MSCMRseg and NCI-ISBI datasets. The Adam optimizer was used with a learning rate of 0.0001, and a consistent batch size of 4 was applied across all datasets. The weighting factors $\lambda_1$, $\lambda_2$, $\lambda_3$, and $\lambda_4$ were empirically set to 0.5, 0.1, 0.1, and 0.1, respectively. The Dice score was employed as the evaluation metric across all datasets.

\subsection{Comparing to SOTA Methods} \label{vs sota}

We compared our MaCo with several advanced WSSS methods and fully supervised methods on the three datasets. Fully supervised methods were trained on masks with pixel-wise annotations, while the WSSS methods included UNet$^+_{pCE}$, UNet$_{RW}$ \cite{grady2006random}, USTM \cite{liu2022weakly}, DMPLS \cite{luo2022scribble}, SC-Net \cite{zhou2023weakly}, S2L \cite{lee2020scribble2label}, CycleMix$_S$ \cite{zhang2022cyclemix}, ScribFormer \cite{li2024scribformer}, nnUNet$_{pL}$ \cite{gotkowski2024embarrassingly}, ShapePU \cite{zhang2022shapepu}, and ScribbleVC \cite{li2023scribblevc}. The fully supervised methods included UNet \cite{baumgartner2018exploration}, CycleMix$_F$ \cite{zhang2022cyclemix}, Swin-UNet \cite{cao2022swin}, TransUNet \cite{chen2024transunet}, Mamba-UNet \cite{wang2024mamba} and MaCo$_F$. The results of these methods were shown in Table \ref{tab:three data}. Among them, the results of CycleMix$_S$, ScribFormer, ShapePU, and ScribbleVC on the ACDC and MSCMRseg datasets were sourced from their respective original papers. Moreover, the results of UNet and CycleMix$_{F}$ on the ACDC and MSCMRseg datasets were taken from the CycleMix paper \cite{zhang2022cyclemix}.

\begin{table}[ht]
  \centering
  \setlength\tabcolsep{10pt}    
  \renewcommand{\arraystretch}{1.2} 
  \caption{Results of MaCo and 10 competing WSSS methods on the ACDC dataset with 70 training samples. Best in each column is \textbf{bold}.}
  \label{tab:70_acdc}

  \resizebox{\linewidth}{!}{
    \begin{tabular}{l  c c c c}
      \toprule
      Method & LV & MYO & RV & Avg. \\
      \midrule
      UNet$_\mathrm{RW}$ \cite{grady2006random}    
        & 87.2 & 71.6 & 80.8 & 79.9 \\
      ShapePU \cite{zhang2022shapepu}              
        & 86.0 & 81.3 & 85.4 & 84.2 \\
      USTM \cite{liu2022weakly}                    
        & 86.3 & 79.8 & 84.0 & 83.4 \\
      nnUNet$_\mathrm{pL}$ \cite{gotkowski2024embarrassingly} 
        & 88.8 & 82.0 & 84.6 & 85.1 \\
      S2L \cite{lee2020scribble2label}             
        & 85.6 & 84.2 & 85.8 & 85.2 \\
      CycleMix$_S$ \cite{zhang2022cyclemix}        
        & 88.0 & 82.5 & 86.0 & 85.5 \\
      DMPLS \cite{luo2022scribble}                 
        & 92.0 & 86.1 & 86.6 & 88.2 \\
      ScribbleVC \cite{li2023scribblevc}           
        & 92.1 & 87.1 & 87.5 & 88.9 \\
      SC-Net \cite{zhou2023weakly}                 
        & 91.9 & 87.5 & 87.9 & 89.1 \\
      ScribFormer \cite{li2024scribformer}         
        & 92.6 & 87.7 & 87.8 & 89.4 \\
      \midrule
      \textbf{MaCo}                              
        & \textbf{93.5} & \textbf{89.5} & \textbf{90.7} & \textbf{91.2} \\
      \bottomrule
    \end{tabular}%
  }
\end{table}

Compared to other scribble-based methods, MaCo outperforms them in most foreground categories across all datasets, except for the MYO segmentation on the MSCMRseg dataset, where it achieves second-best performance. 
The LV, MYO, and RV are spatially adjacent and anatomically intricate structures, making their simultaneous segmentation particularly challenging, especially under sparse scribble supervision. The MYO, in particular, is a thin, ring-shaped structure requiring precise boundary delineation, and is therefore more susceptible to minor segmentation errors. Our analysis indicates that nnUNet$_{pL}$ slightly outperforms MaCo on MYO (by 0.7\%), likely due to its architectural bias toward this category, which appears to reduce its performance on LV and RV. In contrast, MaCo’s design (guided by the ``\textit{from few to more}” principle) promotes a balanced learning strategy, allowing it to generalize effectively across all structures. As a result, MaCo achieves a higher average score on the MSCMRseg dataset, outperforming nnUNet$_{pL}$ by 1.2\%.
Notably, MaCo demonstrates significant improvements of 1.6\%, 1.2\%, and 3.1\%, respectively, across the three datasets, confirming its superior performance.
MaCo$_{F}$ also surpasses five state-of-the-art fully supervised baselines across various architectures: UNet and CycleMix$_{F}$ (CNN-based), Swin-UNet (Transformer-based), Mamba-UNet (Mamba-based), and TransUNet (CNN-Transformer hybrid). Notably, even when trained with only scribble annotations, MaCo slightly outperforms these fully supervised models, highlighting its strong generalization and effective use of weak supervision.

\begin{figure*}[ht]
    \centering
    \includegraphics[width=0.95\linewidth]{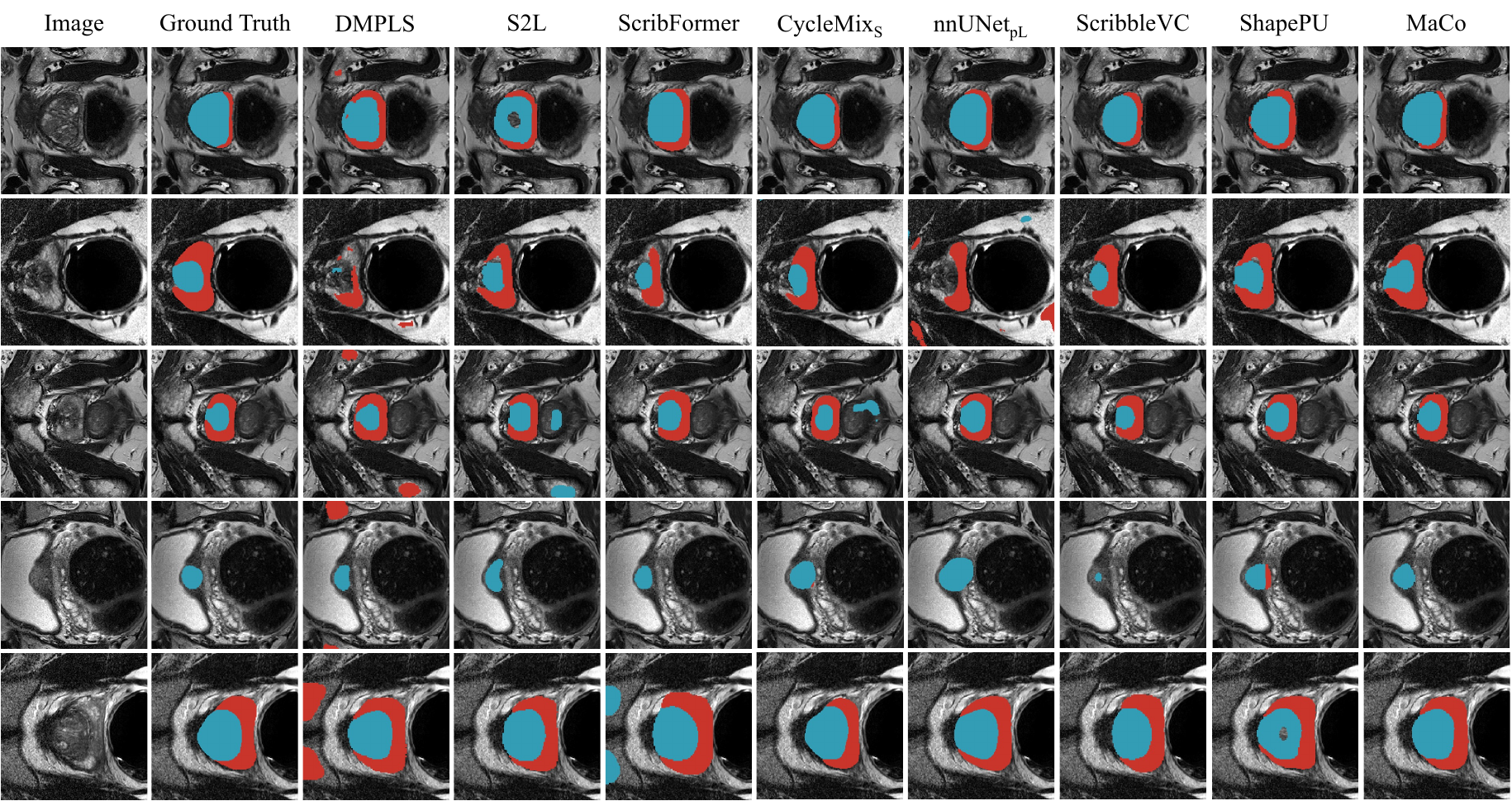}
    \caption{Visualization of segmentation results obtained from seven WSSS methods and MaCo on the NCI-ISBI dataset. Blue and Red are used to color CG and PZ, respectively.}
    \label{fig:nci_result}
\end{figure*}

\subsection{Comparison with Different Masking Strategies}

We evaluated the effectiveness of our structure-aware MCM strategy by comparing it with three commonly used baselines: random sampling \cite{he2022masked}, grid-wise sampling, and uniform $2\times2$ block sampling \cite{li2022uniform} on the ACDC dataset. All strategies were tested using mask ratios of 25\%, 50\%, and 75\%. Fig. \ref{fig:four_mask}(a) shows example masks produced by each method at 25\% masking, while Fig. \ref{fig:four_mask}(b) reports segmentation accuracy under varying mask ratios. Among the baselines, grid mask shows the weakest performance and minimal sensitivity to the masking level. In contrast, our attention-guided masking achieves the best performance, with peak performance at a 50\% masking ratio. These results suggest that the proposed strategy is most effective at extracting target-specific semantic information.

\subsection{Comparison of Different Decay Strategies}

In the CPL module, we adopt an exponential decay function to assign confidence scores based on the distance from each pixel to the nearest scribble annotation. This choice is guided by the principle that confidence should decrease continuously with distance, dropping rapidly near scribbles and tapering more gradually at greater distances. To validate this design, we compare three decay strategies: exponential, Gaussian, and linear. Fig. \ref{fig:three_delay}(a) visualizes their behavior as a function of distance, demonstrating that exponential decay exhibits a faster value reduction at small distances compared to Gaussian decay. Furthermore, by visualizing pseudo-label maps generated with these functions on the ACDC dataset (Fig. \ref{fig:three_delay}(b)), it becomes evident that exponential decay yields a sharper confidence reduction near scribbles, aligning precisely with our desired characteristics for robust pseudo-label generation. Quantitatively, as shown in Table \ref{tab:three_delay}, the exponential decay consistently outperforms both linear and Gaussian decay across all metrics. In summary, the exponential decay function provides an optimal confidence attenuation pattern and achieves superior segmentation performance, justifying its adoption as our decay strategy.

\subsection{Ablation Studies}

We conducted ablation studies to evaluate the effectiveness of the four proposed loss functions: $\mathcal{L}_{\text{mcm}}$, $\mathcal{L}_{\text{mpCE}}$, $\mathcal{L}_{\text{ep}}$, and $\mathcal{L}_{\text{con}}$. The results were presented in Table \ref{tab:ablation_three}. The first row of the table represents the baseline model, trained exclusively with the available scribble annotations.
Our findings can be summarized as follows:
First, the baseline with $\mathcal{L}_{\text{mcm}}$ (\textit{i.e.}, the proposed MCM) significantly outperforms the baseline across all segmentation metrics, showing improvements of 10.5\%, 7.9\%, and 17.1\% for LV, MYO, and RV on the ACDC dataset, respectively. Notably, the average Dice scores on the MSCMRseg and NCI-ISBI datasets increased by 18.1\% and 39.5\%, respectively.
Second, incorporating $\mathcal{L}_{\text{mpCE}}$ and $\mathcal{L}_{\text{ep}}$ leads to further performance gains, with an additional 1.6\%, 7.1\%, and 15.5\% increase in the average Dice score across the three datasets.
Third, the baseline with $\mathcal{L}_{\text{con}}$ (\textit{i.e.}, the proposed CPL) also outperforms the baseline across all metrics, showing an average gain of 8.3\%, 14.5\%, and 23.3\% in Dice scores.
Fourth, when combining all four loss functions, MaCo achieves the highest Dice scores across all three datasets, demonstrating the overall effectiveness of these loss functions in improving the performance of weakly supervised segmentation models.

\begin{table}[t]
  \centering
  \scriptsize                       
  \setlength\tabcolsep{10pt}         
  \caption{Results of the attention mask strategies with different patch sizes on the ACDC dataset. Best in each column is \textbf{bold}.}

  \resizebox{\linewidth}{!}{%
    \begin{tabular}{c  c c c c}
      \toprule
      Patch Size & LV & MYO & RV & Avg. \\
      \midrule
      $4\times4$   & 92.1 & 87.8 & 86.6 & 88.8 \\
      $8\times8$   & 92.6 & 88.8 & 88.5 & 90.0 \\
      $16\times16$ & \textbf{93.4} & \textbf{89.2} & \textbf{88.7} & \textbf{90.4} \\
      $24\times24$ & 92.8 & 89.1 & 88.2 & 90.0 \\
      $32\times32$ & 92.7 & 89.1 & 87.9 & 89.9 \\
      \bottomrule
    \end{tabular}%
  }
  \label{tab:patch_size}
\end{table}

\subsection{More training samples on ACDC dataset}

In Section \ref{vs sota},  we compared MaCo with other advanced WSSS methods using 35 training samples from the ACDC dataset, consistent with common benchmarks. To further assess the effectiveness of MaCo, we conducted additional experiments using a larger set of 70 training samples (encompassing the entire training set of the ACDC dataset). The competing methods included UNet$_{RW}$ \cite{grady2006random}, ShapePU \cite{zhang2022shapepu}, USTM \cite{liu2022weakly}, nnUNet$_{pL}$ \cite{gotkowski2024embarrassingly}, S2L \cite{lee2020scribble2label}, CycleMix$_S$ \cite{zhang2022cyclemix}, DMPLS \cite{luo2022scribble}, ScribbleVC \cite{li2023scribblevc}, SC-Net \cite{zhou2023weakly}, and ScribFormer \cite{li2024scribformer}. The results were shown in Table \ref{tab:70_acdc}. Note that the results of CycleMix$_S$ \cite{zhang2022cyclemix} and ScribFormer \cite{li2024scribformer} were sourced from their respective original papers. 
It reveals that MaCo consistently outperforms all competing methods across all categories. Specifically, MaCo improves Dice scores by 0.9\% for LV, 1.8\% for MYO, 2.8\% for RV, and 1.8\% on average. These results further confirm the superior performance of MaCo with more training samples.

\subsection{Effect of the Decay Factor in CPL}

In Eq.~\ref{eq: cpl}, we adopt an exponential decay function, $e^{-\alpha \cdot D^c(p)}$, to convert distance maps into continuous confidence maps. This section investigates the influence of the decay factor $\alpha$, with values set to 0.05, 0.1, 0.2, and 0.5. The corresponding visualizations and Dice scores are presented in Fig.~\ref{fig:diff_e}(a) and Fig.~\ref{fig:diff_e}(b), respectively. Smaller decay factors (e.g., $\alpha=0.05$) produce broader confidence regions, which enhance supervision by covering more pixels but may introduce greater noise. Conversely, larger values (e.g., $\alpha=0.5$) result in highly localized confidence regions, limiting supervision to areas very close to the scribble annotations. In this extreme case, CPL effectively degenerates into sparse hard labels. Among the tested settings, $\alpha=0.1$ achieves the best Dice scores for LV and MYO and the second-best for RV, yielding the highest overall average. Based on this observation, we select $\alpha=0.1$ as the default setting for CPL.

\subsection{Effect of Hyper-parameter Settings in MCM}

\begin{figure}[ht]
    \centering
    \includegraphics[width=\linewidth]{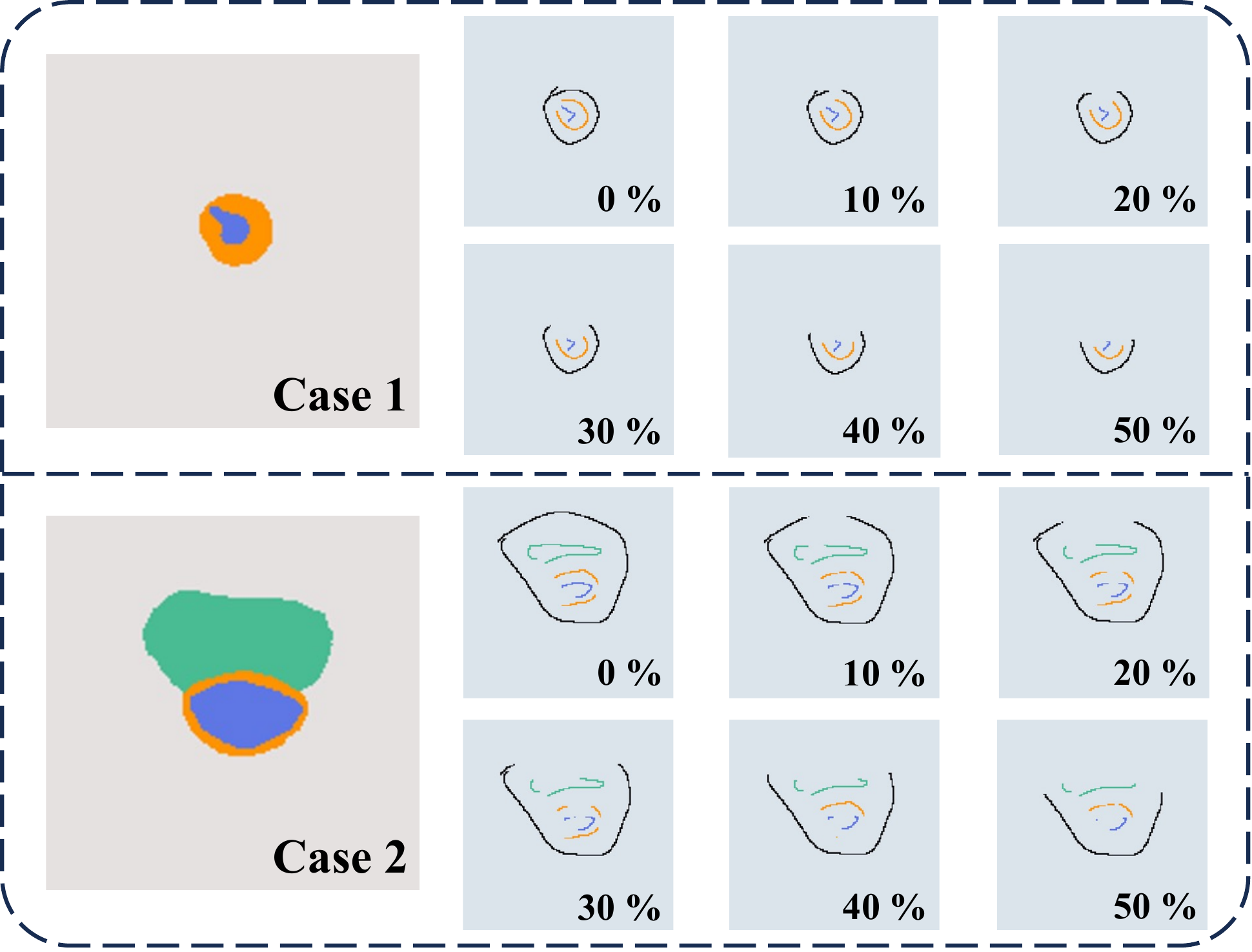}
    \caption{Visualization of scribble annotations for two cases with shrink ratios of 10\%, 20\%, 30\%, 40\%, and 50\%.}
    \label{fig:shrink}
\end{figure}

In the proposed MaCo model, three key hyper-parameters influence segmentation performance: the weight $w_{s}$ for patches containing scribble pixels, the masking ratio $\varphi$, and the patch size used in MCM. We assessed the impact of these hyper-parameters through experiments on the ACDC dataset.

\begin{table}[ht]
  \centering
  \small
  \setlength\tabcolsep{14pt}
  \caption{Performance of MaCo with different shrink ratios and numbers of training samples on the ACDC dataset. Shrink ratio indicates the percentage of masked pixels per category.}
  \begin{tabular}{c cccc}
    \toprule
    \multicolumn{5}{l}{(a) Effect of Shrink Ratio (\%)} \\
    \cmidrule(lr){1-5}
    Ratio & LV & MYO & RV & Avg. \\
    \midrule
    0   & 93.4 & 89.2 & 88.7 & 90.4 \\
    10  & 93.1 & 89.2 & 87.8 & 90.0 \\
    20  & 92.1 & 88.6 & 86.8 & 89.2 \\
    30  & 92.3 & 88.8 & 85.2 & 88.8 \\
    40  & 92.0 & 88.3 & 84.3 & 88.2 \\
    50  & 91.5 & 87.9 & 82.7 & 87.4 \\
    \midrule
    \multicolumn{5}{l}{(b) Effect of Training Sample Number} \\
    \cmidrule(lr){1-5}
    Samples & LV & MYO & RV & Avg. \\
    \midrule
    14  & 90.1 & 85.7 & 82.4 & 86.1 \\
    28  & 90.8 & 87.1 & 87.2 & 88.4 \\
    35  & 93.4 & 89.2 & 88.7 & 90.4 \\
    56  & 93.3 & 89.6 & 88.9 & 90.6 \\
    70  & 93.5 & 89.5 & 90.7 & 91.2 \\
    \bottomrule
  \end{tabular}
  \label{tab:acdc_sensitivity}
\end{table}

We first examined how different combinations of $w_s$ and $\varphi$ impact performance. Specifically, we set the ratio between $w_s$ and $w_o$ (the weight for non-scribble patches) to 1, 2, and 3. When $w_s/w_o$ = 1, the masking strategy simplifies to uniform random masking. For each setting, we varied $\varphi$ from 10\% to 90\% and reported the Dice scores in Fig.~\ref{fig:line_chart}. Across all weight settings, performance initially improves with increasing $\varphi$, peaking at intermediate values and then declining. A larger $w_s/w_o$ leads to more frequent masking of scribble-containing patches, promoting semantic reasoning and improving generalization, particularly when $\varphi$ is low. However, at high masking ratios (e.g., 90\%), this aggressive masking strategy obscures nearly all annotated regions, degrading performance. In such cases, lower $w_s/w_o$ values help preserve some scribbled regions, offering minimal supervision even under heavy occlusion. Overall, a balanced configuration of $w_s/w_o$ = 2 and $\varphi$ = 50\% offers the best trade-off between contextual reasoning and direct supervision, achieving superior segmentation performance.
Next, we tested the effect of patch size by evaluating various dimensions: $4\times4$, $8\times8$, $16\times16$, $24\times24$, and $32\times32$. For dimensions that did not divide the image evenly, zero-padding was applied before patch division. The results, summarized in Table \ref{tab:patch_size}, indicate that a patch size of $16\times16$ provides the best performance.

Based on these findings, the settings adopted for our experiments were $w_{s}=2$, $\varphi=50\%$, and a patch size of $16\times16$.

\subsection{Visualization of Segmentation Results}
For qualitative analysis, we selected five images from the NCI-ISBI dataset and compared the segmentation results from seven WSSS methods and our MaCo, with the images and ground truths, as shown in Fig. \ref{fig:nci_result}. These visualizations demonstrate that the segmentation outputs of MaCo most closely align with the ground truths compared to the other WSSS methods. For example, in the last row, MaCo’s results exhibit the best alignment in shape and size with the ground truth, effectively addressing target omission and overfitting issues observed in other methods.

\begin{figure*}[ht]
    \centering
    \includegraphics[width=\linewidth]{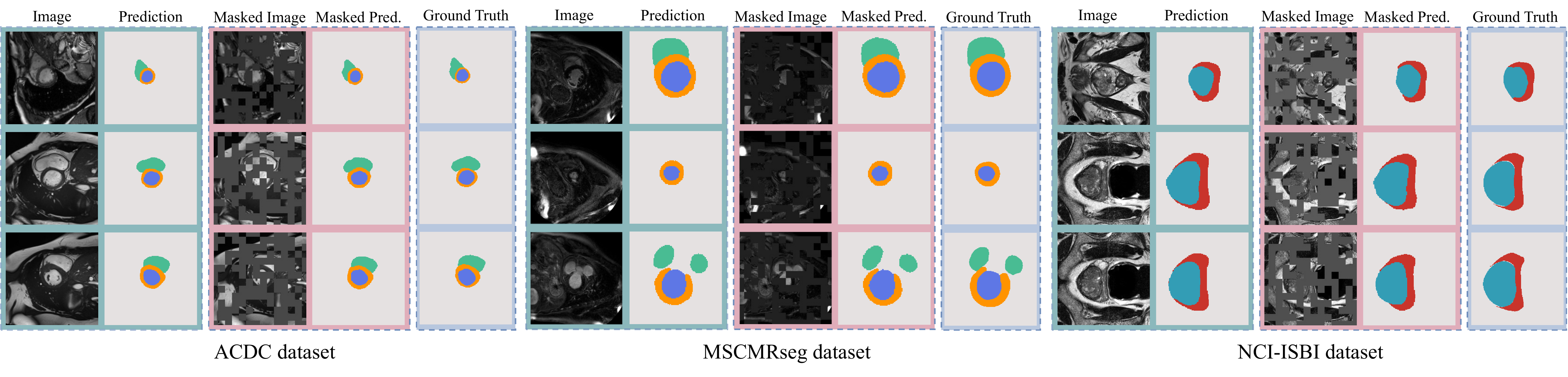}
    \caption{A comparative visualization of masked predictions, original image predictions, and ground truths using MaCo, demonstrated respectively on the ACDC dataset, MSCMRseg dataset, and NCI-ISBI dataset.}
    \label{fig:mask_vs_ori}
\end{figure*}

\section{Discussion}

\subsection{Annotation Sensitivity}

To assess the impact of annotation quality and distribution, we simulate degraded supervision by reducing the length of scribbles using predefined shrink ratios. Specifically, we calculate the target number of pixels to remove from the original scribbles based on the given ratio and apply the following masking protocol to each connected scribble region. For categories containing a single scribble within an image (Fig. \ref{fig:shrink}, Case 1), we progressively mask pixels from one endpoint of the scribble until the specified reduction is achieved. For categories with multiple scribbles (Fig. \ref{fig:shrink}, Case 2), we iteratively remove entire scribbles if doing so does not exceed the target pixel budget. If the remaining budget is smaller than the size of the next full scribble, we partially mask it by removing a contiguous segment to reach the exact reduction threshold. This masking strategy systematically alters both the spatial coverage and density of the annotations, enabling us to evaluate the model’s robustness to varying levels of supervision sparsity and distribution quality.

We visualized the masked scribble labels at shrink ratios of 10\%, 20\%, 30\%, 40\%, and 50\%, as shown in Fig. \ref{fig:shrink}. Unlike random masking, our approach sequentially masks scribbles to better mimic the actual annotation process, where an annotator incrementally adds scribbles rather than simply increasing pixel count. By reducing the number of scribble pixels in each image, we evaluated MaCo's robustness and sensitivity to varying amounts of supervision.
Table \ref{tab:acdc_sensitivity} (a) shows that with 10\% of the scribbles masked, MaCo’s performance drops by only 0.4\%. Even with 50\% of the scribble pixels masked, MaCo retains approximately 96.7\% of its performance with full scribbles.
The observed degradation is attributed to three main factors: (1) reduced supervision strength due to fewer annotated pixels; (2) lower effectiveness of the MCM module, as fewer high-weight patches are available for targeted masking; and (3) diminished accuracy of CPL’s distance-based pseudo labels due to narrower scribble coverage. This controlled degradation emulates real-world low-quality annotation scenarios and demonstrates MaCo’s robustness under such conditions.

\begin{table}[t]
  \centering
  \Large                        
  \setlength\tabcolsep{2pt}      
  \renewcommand{\arraystretch}{1.2}
  \caption{Boundary segmentation accuracy of MaCo and other WSSS methods. The symbol `$-$' means predictions are unavailable. Best in each column is \textbf{bold}.}
  \resizebox{\linewidth}{!}{%
    \begin{tabular}{l  cccc  cccc  ccc}
      \toprule
      \multirow{2}{*}{Method} 
        & \multicolumn{4}{c}{ACDC} 
        & \multicolumn{4}{c}{MSCMRseg} 
        & \multicolumn{3}{c}{NCI-ISBI} \\
      \cmidrule(lr){2-5}\cmidrule(lr){6-9}\cmidrule(l){10-12}
        & LV & MYO & RV & Avg.
        & LV & MYO & RV & Avg.
        & PZ & CG & Avg. \\
      \midrule
      USTM            
        & 82.8 & 77.3 & 76.1 & 78.7 
        & 68.8 & 64.8 & 60.5 & 64.7 
        & 47.8 & 43.4 & 45.6 \\
      SC-Net          
        & 82.6 & 80.2 & 78.3 & 80.4 
        & 67.6 & \textbf{65.4} & \textbf{65.0} & 66.0 
        & 49.9 & 43.8 & 46.9 \\
      UNet$_\mathrm{RW}$ 
        & 72.9 & 63.7 & 63.5 & 66.7 
        & 64.6 & 58.1 & 58.1 & 60.3 
        & 55.0 & 49.2 & 52.1 \\
      ScribFormer     
        & $-$  & $-$  & $-$  & $-$  
        & $-$  & $-$  & $-$  & $-$  
        & 56.6 & 50.5 & 53.6 \\
      CycleMix$_S$    
        & $-$  & $-$  & $-$  & $-$  
        & $-$  & $-$  & $-$  & $-$  
        & 57.1 & 52.3 & 54.7 \\
      ScribbleVC      
        & $-$  & $-$  & $-$  & $-$  
        & $-$  & $-$  & $-$  & $-$  
        & 57.9 & 50.3 & 54.1 \\
      DMPLS           
        & 81.0 & 76.9 & 75.8 & 77.9 
        & 68.9 & 64.9 & 63.2 & 65.7 
        & 60.6 & 50.8 & 55.7 \\
      S2L             
        & 81.6 & 78.0 & 73.2 & 77.6 
        & 70.0 & 64.5 & 64.8 & 66.4 
        & 62.0 & 55.2 & 58.6 \\
      ShapePU         
        & $-$  & $-$  & $-$  & $-$  
        & $-$  & $-$  & $-$  & $-$  
        & \textbf{62.9} & 54.8 & 58.9 \\
      \textbf{MaCo}  
        & \textbf{83.9} & \textbf{81.8} & \textbf{79.8} & \textbf{81.8} 
        & \textbf{72.4} & 65.3 & 63.1 & \textbf{66.9} 
        & \textbf{62.9} & \textbf{57.1} & \textbf{60.0} \\
      \bottomrule
    \end{tabular}%
  }
  \label{tab:boundary_compare}
\end{table}

To further assess the model’s stability with fewer training samples, we conducted experiments with five different training sample sizes on the ACDC dataset, ranging from 14 to 70 samples. The results, shown in Table \ref{tab:acdc_sensitivity} (b), reveal that as the number of training samples increased from 14 to 70, MaCo’s performance improves from 86.1\% to 91.2\% in terms of the average Dice score, suggesting that MaCo’s segmentation performance can be further enhanced with more scribble annotations.

In summary, MaCo shows a robust ability to learn from sparse annotations for pixel-wise dense predictions, outperforming other state-of-the-art WSSS methods. More important, MaCo significantly reduces the need for pixel-wise annotations while achieving only a small performance decrease with less supervision.

\begin{figure}[t]
    \centering
    \includegraphics[width=0.95\linewidth]{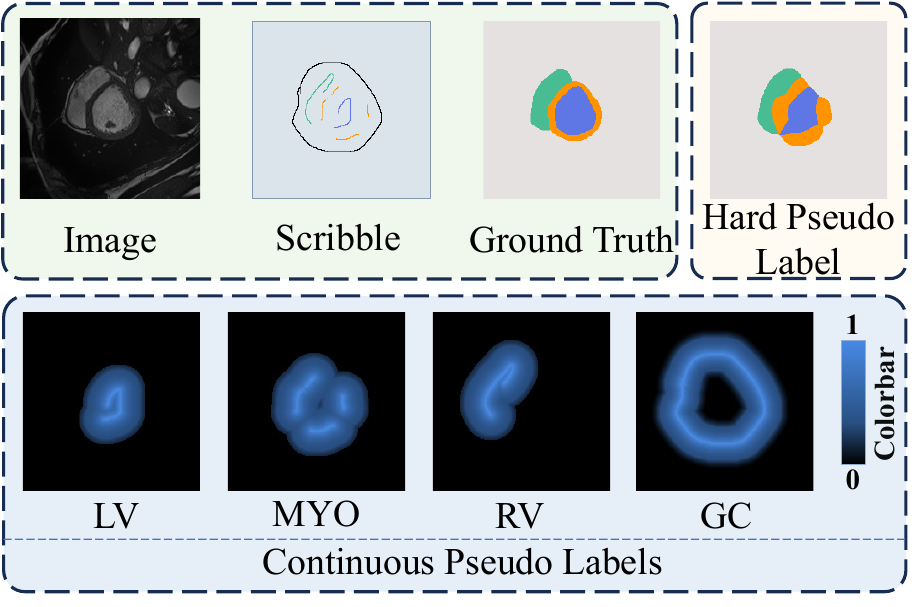}
    \caption{Visualization of two types of pseudo-labels, hard pseudo label and continuous pseudo labels on a sample from the ACDC dataset.}
    \label{fig:hpl_cpl}
\end{figure}

\subsection{Masked Contextual Comprehension}

The proposed MCM component generates masked images from the original input using an attention-based masking operation. Both the original and masked images are then fed into the model to obtain similar predictions. Fig. \ref{fig:mask_vs_ori} illustrates the predictions for both input types. The visualizations reveal a high degree of similarity between the predictions, particularly in the masked regions. This demonstrates MaCo’s ability to infer missing information from contextual cues, effectively extrapolating from partial data. For instance, in the third row of the ACDC dataset, the model’s predictions closely align with the ground truth, even though the majority of the target region is masked in the input image.

\subsection{Boundary Segmentation Accuracy}

Boundary segmentation accuracy remains one of the most significant challenges in segmentation tasks. We compared MaCo with other WSSS methods in terms of boundary extraction accuracy. Specifically, we extracted the boundaries of each foreground category from ground truths and calculated the proportion of boundary pixels correctly identified by the model. The results, presented in Table \ref{tab:boundary_compare}, show that MaCo achieves the highest accuracy across all categories in the ACDC and NCI-ISBI datasets. While the boundary accuracy for MYO and RV in the MSCMRseg dataset is slightly lower, MaCo still outperforms all other methods, achieving the highest overall accuracy. This demonstrates that the proposed CPL generation in MaCo effectively uses confidence levels to create continuous pseudo labels, as opposed to relying on hard pseudo labels, leading to more accurate boundary segmentation.

\subsection{Continuous Pseudo Labels VS. Hard Pseudo Labels}

We acknowledge that all pseudo-labeling approaches, including distance-based hard pseudo labels (HPL) and our CPL, may inevitably introduce misleading signals. In HPL, each unlabeled pixel is assigned the class of its nearest scribble. However, CPL's inherent design effectively mitigates their negative impact on model learning. HPL rigidly assigns uniform, full-confidence weights across entire category regions, often leading to substantial erroneous supervision where boundaries are complex or initial annotations are sparse. In contrast, while CPL, being distance-based, can also introduce errors, especially near intricate anatomical boundaries or under extremely sparse scribble annotations, its design inherently limits their influence. Pixels more prone to error, being further from scribbles, are assigned lower confidence levels, thus contributing less to the training signal. Fig. \ref{fig:hpl_cpl} visualizes this distinction, clearly showing that HPL contains numerous misleading signals, whereas CPL assigns high confidence near scribble annotations that gracefully decays with distance. Furthermore, quantitative experiments in Table \ref{tab:soft_hard} demonstrate that CPL consistently yields significant performance improvements across all datasets compared to HPL, affirming its robustness against misleading signals.

\begin{table}[t]
  \centering
  \Large                        
  \setlength\tabcolsep{2pt}      
  \renewcommand{\arraystretch}{1.2}
  \caption{Performance comparison with hard pseudo labels on three datasets. HPL means Hard Pseudo Labels, CPL means Continuous Pseudo Labels. Best in each column is \textbf{bold}.}
  \resizebox{\linewidth}{!}{
    \begin{tabular}{c  cccc  cccc  ccc}
      \toprule
      \multirow{2}{*}{Method} 
        & \multicolumn{4}{c}{ACDC} 
        & \multicolumn{4}{c}{MSCMRseg} 
        & \multicolumn{3}{c}{NCI-ISBI} \\
      \cmidrule(lr){2-5}\cmidrule(lr){6-9}\cmidrule(l){10-12}
        & LV & MYO & RV & Avg.
        & LV & MYO & RV & Avg.
        & PZ & CG & Avg. \\
      \midrule
      MaCo w/ HPL       
        & 92.4 & 88.7 & 81.1 & 87.4
        & 89.4 & 81.0 & 85.6 & 85.3
        & 56.3 & 75.0 & 65.7  \\
      MaCo w/ CPL
        & \textbf{93.4} & \textbf{89.2} & \textbf{88.7} & \textbf{90.4} 
        & \textbf{93.1} & \textbf{84.7} & \textbf{87.8} & \textbf{88.5} 
        & \textbf{73.8} & \textbf{87.1} & \textbf{80.5} \\
      \bottomrule
    \end{tabular}%
  }
  \label{tab:soft_hard}
\end{table}

\subsection{Potential of Extending MaCo to Natural Domain}

The core principle of MaCo, \textit{from few to more}, is broadly applicable to weakly supervised segmentation tasks, making it a promising candidate for extension beyond medical imaging to natural image segmentation. Both domains share the central challenge of inferring dense, pixel-level labels from sparse annotations. However, applying MaCo to natural images introduces unique challenges. A primary concern is the significantly larger number of semantic categories typically present in natural image datasets, such as the 21 in ScribbleSup \cite{lin2016scribblesup} and 81 in ScribbleCOCO \cite{zhang2025exploiting}. This high class diversity increases the likelihood that a single masked region contains pixels from multiple categories, complicating the pretext task of MCM. As a result, key hyperparameters such as the masking ratio and patch size would need to be carefully adjusted to preserve semantic coherence and ensure effective learning. To address these challenges, future work could investigate adaptive or class-aware masking strategies that dynamically adjust to the local content and category distribution of natural scenes.

\begin{table}[th]
  \centering
  \large                     
  \setlength\tabcolsep{5pt} 
  \renewcommand{\arraystretch}{1.2} 
  \caption{Comparison of Training and Inference Efficiency for the baseline, SD, and MCM  on the ACDC dataset. Baseline: Using an original image for training.  SD: Using self-distillation learning for training. \#Params.: Number of parameters; GPU Mem.: GPU Memory cost when training; Tra. Time: Training Time each epoch; Inf. Time: Inference Time each image.}
  \resizebox{\linewidth}{!}{
    \begin{tabular}{l  ccccc}
      \toprule
      Method & \#Params (M) & GPU Mem. (MB) & Tra. Time  (s) & Inf. Time (s) & Dice (\%) \\
      \midrule
      Baseline     & 25.76 & 2665.28 & 40 & \num{1.15e-2} & 86.5 \\
      SD           & 25.76 & 5046.86 & 67 & \num{1.15e-2} & 87.6 \\
      Ours         & 25.76 & 5074.55 & 69 & \num{1.15e-2} & 90.4 \\
      
      \bottomrule
    \end{tabular}%
  }
  \label{tab:test_time_MCM}
\end{table}

\begin{table}[th]
  \centering
  \large                     
  \setlength\tabcolsep{5pt} 
  \renewcommand{\arraystretch}{1.2} 
  \caption{Costs of generating continuous pseudo labels by CPL on three datasets. \#training images: Number of training images; \#classes.: Number of classes. Total time means time cost of generating all continuous pseudo labels; w/wo CPL (s) means the training time cost  per epoch with or without CPL.}
  \resizebox{\linewidth}{!}{
    \begin{tabular}{l  ccccc}
      \toprule
      Dataset & \#training images & Total time (s) & \#classes & Average shape & w/wo CPL (s)
\\
      \midrule
      ACDC     & 656 & 13 & 4 & (212, 246) & 69/66 \\
      MSCMRseg & 382 & 29 & 4 & (481, 481) & 41/39 \\
      NCI-ISBI & 807 & 37 & 3 & (363, 363) & 85/82 \\
      
      \bottomrule
    \end{tabular}%
  }
  \label{tab:test_time_CPL}
\end{table}

\subsection{Resource Requirements for Training and Inference}

Our MaCo framework introduces two components, MCM and CPL, to significantly enhance WSSS performance. This section evaluates their computational costs during training and inference. 
MCM operates online during training and introduces additional overhead due to a second forward pass for masked images. This is efficiently implemented by concatenating the original and masked images along the batch dimension. We compare MCM with a baseline (training with only the original image) and a self-distillation (SD) approach based on BYOL \cite{grill2020bootstrap}, which generates two augmented views and enforces output consistency. As shown in Table \ref{tab:test_time_MCM}, (1) neither MCM nor SD methods increase model parameters or inference time, since the additional inputs are used only during training; (2) both MCM and SD methods approximately double GPU memory consumption during training and increase training time per epoch by approximately 1.75$\times$ due to the enlarged effective batch size; and (3) these costs are justified by substantial performance gains: MCM improves Dice by 3.9\% over the baseline and 2.8\% over SD on the ACDC dataset. CPL operates offline before training and thus incurs negligible runtime cost. As reported in Table \ref{tab:test_time_CPL}, generating continuous pseudo labels for the ACDC, MSCMRseg, and NCI-ISBI datasets requires only 13, 29, and 37 seconds, respectively. Incorporating CPL into training adds just 2–3 seconds per epoch across all datasets (e.g., 3s for ACDC, 2s for MSCMRseg, 3s for NCI-ISBI), while significantly enhancing segmentation performance. 

In summary, MCM and CPL introduce modest training-time overhead but do not impact inference efficiency. Their computational demands are acceptable and scalable, especially in light of the substantial accuracy improvements they enable.

\section{Conclusion}

In this paper, we propose MaCo, a weakly supervised model for medical image segmentation that utilizes MCM and CPL. Following the ``\textit{from few to more}" principle, MCM enhances the model's ability to align predictions from both the original and masked versions of an image, improving its capacity to infer semantic information from contextual cues. CPL generates continuous pseudo labels, which strike a balance between providing rich supervisory signals and addressing potential ambiguities in the data. Our experiments on three scribble-based segmentation datasets demonstrate the effectiveness of MaCo in improving segmentation performance. 
In future work, we aim to further explore techniques to refine and expand the model's ability to learn from sparse supervision, enhancing its robustness and accuracy in medical image segmentation tasks.

\section*{References}

\bibliographystyle{IEEEtran}
\bibliography{MaCo_references}

\begin{thebibliography}{10}
\providecommand{\url}[1]{#1}
\csname url@samestyle\endcsname
\providecommand{\newblock}{\relax}
\providecommand{\bibinfo}[2]{#2}
\providecommand{\BIBentrySTDinterwordspacing}{\spaceskip=0pt\relax}
\providecommand{\BIBentryALTinterwordstretchfactor}{4}
\providecommand{\BIBentryALTinterwordspacing}{\spaceskip=\fontdimen2\font plus
\BIBentryALTinterwordstretchfactor\fontdimen3\font minus \fontdimen4\font\relax}
\providecommand{\BIBforeignlanguage}[2]{{%
\expandafter\ifx\csname l@#1\endcsname\relax
\typeout{** WARNING: IEEEtran.bst: No hyphenation pattern has been}%
\typeout{** loaded for the language `#1'. Using the pattern for}%
\typeout{** the default language instead.}%
\else
\language=\csname l@#1\endcsname
\fi
#2}}
\providecommand{\BIBdecl}{\relax}
\BIBdecl

\bibitem{azad2024medical}
R.~Azad, E.~K. Aghdam, A.~Rauland, Y.~Jia, A.~H. Avval, A.~Bozorgpour, S.~Karimijafarbigloo, J.~P. Cohen, E.~Adeli, and D.~Merhof, ``Medical image segmentation review: The success of u-net,'' \emph{IEEE Transactions on Pattern Analysis and Machine Intelligence}, 2024.

\bibitem{panayides2020ai}
A.~S. Panayides, A.~Amini, N.~D. Filipovic, A.~Sharma, S.~A. Tsaftaris, A.~Young, D.~Foran, N.~Do, S.~Golemati, T.~Kurc \emph{et~al.}, ``Ai in medical imaging informatics: current challenges and future directions,'' \emph{IEEE journal of biomedical and health informatics}, vol.~24, no.~7, pp. 1837--1857, 2020.

\bibitem{liu2021review}
X.~Liu, L.~Song, S.~Liu, and Y.~Zhang, ``A review of deep-learning-based medical image segmentation methods,'' \emph{Sustainability}, vol.~13, no.~3, p. 1224, 2021.

\bibitem{jin2023label}
C.~Jin, Z.~Guo, Y.~Lin, L.~Luo, and H.~Chen, ``Label-efficient deep learning in medical image analysis: Challenges and future directions,'' \emph{arXiv preprint arXiv:2303.12484}, 2023.

\bibitem{tajbakhsh2020embracing}
N.~Tajbakhsh, L.~Jeyaseelan, Q.~Li, J.~N. Chiang, Z.~Wu, and X.~Ding, ``Embracing imperfect datasets: A review of deep learning solutions for medical image segmentation,'' \emph{Medical image analysis}, vol.~63, p. 101693, 2020.

\bibitem{zhou2003learning}
D.~Zhou, O.~Bousquet, T.~Lal, J.~Weston, and B.~Sch{\"o}lkopf, ``Learning with local and global consistency,'' \emph{Advances in neural information processing systems}, vol.~16, 2003.

\bibitem{luo2021semi}
X.~Luo, J.~Chen, T.~Song, and G.~Wang, ``Semi-supervised medical image segmentation through dual-task consistency,'' in \emph{Proceedings of the AAAI conference on artificial intelligence}, vol.~35, no.~10, 2021, pp. 8801--8809.

\bibitem{zhang2022shapepu}
K.~Zhang and X.~Zhuang, ``Shapepu: A new pu learning framework regularized by global consistency for scribble supervised cardiac segmentation,'' in \emph{International Conference on Medical Image Computing and Computer-Assisted Intervention}.\hskip 1em plus 0.5em minus 0.4em\relax Springer, 2022, pp. 162--172.

\bibitem{zhang2022cyclemix}
------, ``Cyclemix: A holistic strategy for medical image segmentation from scribble supervision,'' in \emph{Proceedings of the IEEE/CVF Conference on Computer Vision and Pattern Recognition}, 2022, pp. 11\,656--11\,665.

\bibitem{lee2020scribble2label}
H.~Lee and W.-K. Jeong, ``Scribble2label: Scribble-supervised cell segmentation via self-generating pseudo-labels with consistency,'' in \emph{Medical Image Computing and Computer Assisted Intervention--MICCAI 2020: 23rd International Conference, Lima, Peru, October 4--8, 2020, Proceedings, Part I 23}.\hskip 1em plus 0.5em minus 0.4em\relax Springer, 2020, pp. 14--23.

\bibitem{zhou2023weakly}
M.~Zhou, Z.~Xu, K.~Zhou, and R.~K.-y. Tong, ``Weakly supervised medical image segmentation via superpixel-guided scribble walking and class-wise contrastive regularization,'' in \emph{International Conference on Medical Image Computing and Computer-Assisted Intervention}.\hskip 1em plus 0.5em minus 0.4em\relax Springer, 2023, pp. 137--147.

\bibitem{luo2022scribble}
X.~Luo, M.~Hu, W.~Liao, S.~Zhai, T.~Song, G.~Wang, and S.~Zhang, ``Scribble-supervised medical image segmentation via dual-branch network and dynamically mixed pseudo labels supervision,'' in \emph{International Conference on Medical Image Computing and Computer-Assisted Intervention}.\hskip 1em plus 0.5em minus 0.4em\relax Springer, 2022, pp. 528--538.

\bibitem{li2023scribblevc}
Z.~Li, Y.~Zheng, X.~Luo, D.~Shan, and Q.~Hong, ``Scribblevc: Scribble-supervised medical image segmentation with vision-class embedding,'' in \emph{Proceedings of the 31st ACM International Conference on Multimedia}, 2023, pp. 3384--3393.

\bibitem{li2024scribformer}
Z.~Li, Y.~Zheng, D.~Shan, S.~Yang, Q.~Li, B.~Wang, Y.~Zhang, Q.~Hong, and D.~Shen, ``Scribformer: Transformer makes cnn work better for scribble-based medical image segmentation,'' \emph{IEEE Transactions on Medical Imaging}, 2024.

\bibitem{shen2023survey}
W.~Shen, Z.~Peng, X.~Wang, H.~Wang, J.~Cen, D.~Jiang, L.~Xie, X.~Yang, and Q.~Tian, ``A survey on label-efficient deep image segmentation: Bridging the gap between weak supervision and dense prediction,'' \emph{IEEE transactions on pattern analysis and machine intelligence}, vol.~45, no.~8, pp. 9284--9305, 2023.

\bibitem{han2022multi}
C.~Han, J.~Lin, J.~Mai, Y.~Wang, Q.~Zhang, B.~Zhao, X.~Chen, X.~Pan, Z.~Shi, Z.~Xu \emph{et~al.}, ``Multi-layer pseudo-supervision for histopathology tissue semantic segmentation using patch-level classification labels,'' \emph{Medical Image Analysis}, vol.~80, p. 102487, 2022.

\bibitem{tang2024hunting}
F.~Tang, Z.~Xu, Z.~Qu, W.~Feng, X.~Jiang, and Z.~Ge, ``Hunting attributes: Context prototype-aware learning for weakly supervised semantic segmentation,'' in \emph{Proceedings of the IEEE/CVF Conference on Computer Vision and Pattern Recognition}, 2024, pp. 3324--3334.

\bibitem{zhao2024sfc}
X.~Zhao, F.~Tang, X.~Wang, and J.~Xiao, ``Sfc: Shared feature calibration in weakly supervised semantic segmentation,'' in \emph{Proceedings of the AAAI Conference on Artificial Intelligence}, vol.~38, no.~7, 2024, pp. 7525--7533.

\bibitem{roth2021going}
H.~R. Roth, D.~Yang, Z.~Xu, X.~Wang, and D.~Xu, ``Going to extremes: weakly supervised medical image segmentation,'' \emph{Machine Learning and Knowledge Extraction}, vol.~3, no.~2, pp. 507--524, 2021.

\bibitem{zhong2023simple}
Y.~Zhong and Y.~Wang, ``Simple: Similarity-aware propagation learning for weakly-supervised breast cancer segmentation in dce-mri,'' in \emph{International Conference on Medical Image Computing and Computer-Assisted Intervention}.\hskip 1em plus 0.5em minus 0.4em\relax Springer, 2023, pp. 567--577.

\bibitem{en2022annotation}
Q.~En and Y.~Guo, ``Annotation by clicks: A point-supervised contrastive variance method for medical semantic segmentation,'' \emph{arXiv preprint arXiv:2212.08774}, 2022.

\bibitem{wei2023weakpolyp}
J.~Wei, Y.~Hu, S.~Cui, S.~K. Zhou, and Z.~Li, ``Weakpolyp: You only look bounding box for polyp segmentation,'' in \emph{International Conference on Medical Image Computing and Computer-Assisted Intervention}.\hskip 1em plus 0.5em minus 0.4em\relax Springer, 2023, pp. 757--766.

\bibitem{wang2021bounding}
J.~Wang and B.~Xia, ``Bounding box tightness prior for weakly supervised image segmentation,'' in \emph{International conference on medical image computing and computer-assisted intervention}.\hskip 1em plus 0.5em minus 0.4em\relax Springer, 2021, pp. 526--536.

\bibitem{liu2022weakly}
X.~Liu, Q.~Yuan, Y.~Gao, K.~He, S.~Wang, X.~Tang, J.~Tang, and D.~Shen, ``Weakly supervised segmentation of covid19 infection with scribble annotation on ct images,'' \emph{Pattern recognition}, vol. 122, p. 108341, 2022.

\bibitem{gotkowski2024embarrassingly}
K.~Gotkowski, C.~L{\"u}th, P.~F. J{\"a}ger, S.~Ziegler, L.~Kr{\"a}mer, S.~Denner, S.~Xiao, N.~Disch, K.~H. Maier-Hein, and F.~Isensee, ``Embarrassingly simple scribble supervision for 3d medical segmentation,'' \emph{arXiv preprint arXiv:2403.12834}, 2024.

\bibitem{he2022masked}
K.~He, X.~Chen, S.~Xie, Y.~Li, P.~Doll{\'a}r, and R.~Girshick, ``Masked autoencoders are scalable vision learners,'' in \emph{Proceedings of the IEEE/CVF conference on computer vision and pattern recognition}, 2022, pp. 16\,000--16\,009.

\bibitem{cai2022uni4eye}
Z.~Cai, L.~Lin, H.~He, and X.~Tang, ``Uni4eye: unified 2d and 3d self-supervised pre-training via masked image modeling transformer for ophthalmic image classification,'' in \emph{International Conference on Medical Image Computing and Computer-Assisted Intervention}.\hskip 1em plus 0.5em minus 0.4em\relax Springer, 2022, pp. 88--98.

\bibitem{wang2023swinmm}
Y.~Wang, Z.~Li, J.~Mei, Z.~Wei, L.~Liu, C.~Wang, S.~Sang, A.~L. Yuille, C.~Xie, and Y.~Zhou, ``Swinmm: masked multi-view with swin transformers for 3d medical image segmentation,'' in \emph{International conference on medical image computing and computer-assisted intervention}.\hskip 1em plus 0.5em minus 0.4em\relax Springer, 2023, pp. 486--496.

\bibitem{ye2024continual}
Y.~Ye, Y.~Xie, J.~Zhang, Z.~Chen, Q.~Wu, and Y.~Xia, ``Continual self-supervised learning: Towards universal multi-modal medical data representation learning,'' in \emph{Proceedings of the IEEE/CVF Conference on Computer Vision and Pattern Recognition}, 2024, pp. 11\,114--11\,124.

\bibitem{wang2023imagen}
S.~Wang, C.~Saharia, C.~Montgomery, J.~Pont-Tuset, S.~Noy, S.~Pellegrini, Y.~Onoe, S.~Laszlo, D.~J. Fleet, R.~Soricut \emph{et~al.}, ``Imagen editor and editbench: Advancing and evaluating text-guided image inpainting,'' in \emph{Proceedings of the IEEE/CVF conference on computer vision and pattern recognition}, 2023, pp. 18\,359--18\,369.

\bibitem{chang2022maskgit}
H.~Chang, H.~Zhang, L.~Jiang, C.~Liu, and W.~T. Freeman, ``Maskgit: Masked generative image transformer,'' in \emph{Proceedings of the IEEE/CVF Conference on Computer Vision and Pattern Recognition}, 2022, pp. 11\,315--11\,325.

\bibitem{gao2023masked}
S.~Gao, P.~Zhou, M.-M. Cheng, and S.~Yan, ``Masked diffusion transformer is a strong image synthesizer,'' in \emph{Proceedings of the IEEE/CVF International Conference on Computer Vision}, 2023, pp. 23\,164--23\,173.

\bibitem{yuan2022msml}
G.~Yuan, H.~Zheng, and J.~Dong, ``Msml: Enhancing occlusion-robustness by multi-scale segmentation-based mask learning for face recognition,'' in \emph{Proceedings of the AAAI Conference on Artificial Intelligence}, 2022, pp. 3197--3205.

\bibitem{ji2023multi}
S.~Ji, S.~Han, and J.~Rhee, ``Multi-view masked autoencoder for general image representation,'' \emph{Applied Sciences}, vol.~13, no.~22, p. 12413, 2023.

\bibitem{abdelfattah2024maskclr}
M.~Abdelfattah, M.~Hassan, and A.~Alahi, ``Maskclr: Attention-guided contrastive learning for robust action representation learning,'' in \emph{Proceedings of the IEEE/CVF Conference on Computer Vision and Pattern Recognition}, 2024, pp. 18\,678--18\,687.

\bibitem{baumgartner2018exploration}
C.~F. Baumgartner, L.~M. Koch, M.~Pollefeys, and E.~Konukoglu, ``An exploration of 2d and 3d deep learning techniques for cardiac mr image segmentation,'' in \emph{Statistical Atlases and Computational Models of the Heart. ACDC and MMWHS Challenges: 8th International Workshop, STACOM 2017, Held in Conjunction with MICCAI 2017, Quebec City, Canada, September 10-14, 2017, Revised Selected Papers 8}.\hskip 1em plus 0.5em minus 0.4em\relax Springer, 2018, pp. 111--119.

\bibitem{ronneberger2015u}
O.~Ronneberger, P.~Fischer, and T.~Brox, ``U-net: Convolutional networks for biomedical image segmentation,'' in \emph{Medical image computing and computer-assisted intervention--MICCAI 2015: 18th international conference, Munich, Germany, October 5-9, 2015, proceedings, part III 18}.\hskip 1em plus 0.5em minus 0.4em\relax Springer, 2015, pp. 234--241.

\bibitem{cao2022swin}
H.~Cao, Y.~Wang, J.~Chen, D.~Jiang, X.~Zhang, Q.~Tian, and M.~Wang, ``Swin-unet: Unet-like pure transformer for medical image segmentation,'' in \emph{European conference on computer vision}.\hskip 1em plus 0.5em minus 0.4em\relax Springer, 2022, pp. 205--218.

\bibitem{wang2024mamba}
Z.~Wang, J.-Q. Zheng, Y.~Zhang, G.~Cui, and L.~Li, ``Mamba-unet: Unet-like pure visual mamba for medical image segmentation,'' \emph{arXiv preprint arXiv:2402.05079}, 2024.

\bibitem{chen2024transunet}
J.~Chen, J.~Mei, X.~Li, Y.~Lu, Q.~Yu, Q.~Wei, X.~Luo, Y.~Xie, E.~Adeli, Y.~Wang \emph{et~al.}, ``Transunet: Rethinking the u-net architecture design for medical image segmentation through the lens of transformers,'' \emph{Medical Image Analysis}, p. 103280, 2024.

\bibitem{grady2006random}
L.~Grady, ``Random walks for image segmentation,'' \emph{IEEE transactions on pattern analysis and machine intelligence}, vol.~28, no.~11, pp. 1768--1783, 2006.

\bibitem{tang2018normalized}
M.~Tang, A.~Djelouah, F.~Perazzi, Y.~Boykov, and C.~Schroers, ``Normalized cut loss for weakly-supervised cnn segmentation,'' in \emph{Proceedings of the IEEE conference on computer vision and pattern recognition}, 2018, pp. 1818--1827.

\bibitem{bernard2018deep}
O.~Bernard, A.~Lalande, C.~Zotti, F.~Cervenansky, X.~Yang, P.-A. Heng, I.~Cetin, K.~Lekadir, O.~Camara, M.~A.~G. Ballester \emph{et~al.}, ``Deep learning techniques for automatic mri cardiac multi-structures segmentation and diagnosis: is the problem solved?'' \emph{IEEE transactions on medical imaging}, vol.~37, no.~11, pp. 2514--2525, 2018.

\bibitem{valvano2021learning}
G.~Valvano, A.~Leo, and S.~A. Tsaftaris, ``Learning to segment from scribbles using multi-scale adversarial attention gates,'' \emph{IEEE Transactions on Medical Imaging}, vol.~40, no.~8, pp. 1990--2001, 2021.

\bibitem{zhuang2018multivariate}
X.~Zhuang, ``Multivariate mixture model for myocardial segmentation combining multi-source images,'' \emph{IEEE transactions on pattern analysis and machine intelligence}, vol.~41, no.~12, pp. 2933--2946, 2018.

\bibitem{clark2013cancer}
K.~Clark, B.~Vendt, K.~Smith, J.~Freymann, J.~Kirby, P.~Koppel, S.~Moore, S.~Phillips, D.~Maffitt, M.~Pringle \emph{et~al.}, ``The cancer imaging archive (tcia): maintaining and operating a public information repository,'' \emph{Journal of digital imaging}, vol.~26, pp. 1045--1057, 2013.

\bibitem{li2022uniform}
X.~Li, W.~Wang, L.~Yang, and J.~Yang, ``Uniform masking: Enabling mae pre-training for pyramid-based vision transformers with locality,'' \emph{arXiv preprint arXiv:2205.10063}, 2022.

\bibitem{lin2016scribblesup}
D.~Lin, J.~Dai, J.~Jia, K.~He, and J.~Sun, ``Scribblesup: Scribble-supervised convolutional networks for semantic segmentation,'' in \emph{Proceedings of the IEEE conference on computer vision and pattern recognition}, 2016, pp. 3159--3167.

\bibitem{zhang2025exploiting}
X.~Zhang, L.~Zhu, S.~Zeng, H.~He, O.~Fu, Z.~Yao, Z.~Xie, and Y.~Lu, ``Exploiting inherent class label: Towards robust scribble supervised semantic segmentation,'' \emph{arXiv preprint arXiv:2503.13895}, 2025.

\bibitem{grill2020bootstrap}
J.-B. Grill, F.~Strub, F.~Altch{\'e}, C.~Tallec, P.~Richemond, E.~Buchatskaya, C.~Doersch, B.~Avila~Pires, Z.~Guo, M.~Gheshlaghi~Azar \emph{et~al.}, ``Bootstrap your own latent-a new approach to self-supervised learning,'' \emph{Advances in neural information processing systems}, vol.~33, pp. 21\,271--21\,284, 2020.

\end{thebibliography}

\end{document}